\documentclass[runningheads]{llncs}
\usepackage{rotating}
\usepackage{graphicx}
\usepackage{amsmath,amssymb} 
\usepackage{url}
\usepackage{tikz}
\usepackage{pgfplots}
\pgfplotsset{compat=newest}

\usepackage{color}
\usepackage{pifont}
\usepackage[width=122mm,left=12mm,paperwidth=146mm,height=193mm,top=12mm,paperheight=217mm]{geometry}

\newcommand{\showodsf}[3]{%
\mbox{\input{data/pr/#1_#2_#3_ods_f.txt}\hspace{-2.5pt}}%
}

\usepackage{subfiles}

\usepackage{booktabs}
\usepackage{multicol}
\definecolor{rowblue}{RGB}{220,230,240}

\usepackage{capt-of}

\newcommand{\cfbox}[2]{%
    \colorlet{currentcolor}{.}%
    {\color{#1}%
    \fbox{\color{currentcolor}#2}}%
}

\newcommand{\showonestuff}[6]{%
\fbox{\includegraphics[height=#6\linewidth]{img/supplemental/#1_#5.png}}
\hfill
\fbox{\includegraphics[height=#6\linewidth]{img/supplemental/#2_#5.png}}
\hfill
\fbox{\includegraphics[height=#6\linewidth]{img/supplemental/#3_#5.png}}
\hfill
\fbox{\includegraphics[height=#6\linewidth]{img/supplemental/#4_#5.png}}
\vspace{2mm}
}

\begin{document}
\pagestyle{headings}
\mainmatter

\title{Convolutional Oriented Boundaries} 
\titlerunning{Convolutional Oriented Boundaries}
\authorrunning{K.K. Maninis, J. Pont-Tuset, P. Arbel\'{a}ez, L. Van Gool}

\author{Kevis-Kokitsi Maninis\inst{1},$\quad$Jordi Pont-Tuset\inst{1},\\Pablo Arbel\'{a}ez,\inst{2}$\quad$Luc Van Gool\inst{1,3}}
\institute{$^1$ETH Z\"urich $\quad$ $^2$Universidad de los Andes $\quad$ $^3$KU Leuven}

\maketitle

\begin{abstract}
We present Convolutional Oriented Boundaries (COB), which produces multiscale oriented contours and region hierarchies starting from generic image classification Convolutional Neural Networks (CNNs).
COB is computationally efficient, because it requires a single CNN forward pass for contour detection and it uses a novel sparse boundary representation for hierarchical segmentation; it gives a significant leap in performance over the state-of-the-art, and it generalizes very well to unseen categories and datasets. Particularly, we show that learning to estimate not only contour strength but also orientation provides more accurate results. 
We perform extensive experiments on BSDS, PASCAL Context, PASCAL Segmentation, and MS-COCO, showing that COB provides state-of-the-art contours, region hierarchies, and object proposals in all datasets.
\end{abstract}

\section{Introduction}
\label{sec:intro}

\begin{figure}[t]
\centering
\includegraphics[width=0.9\textwidth]{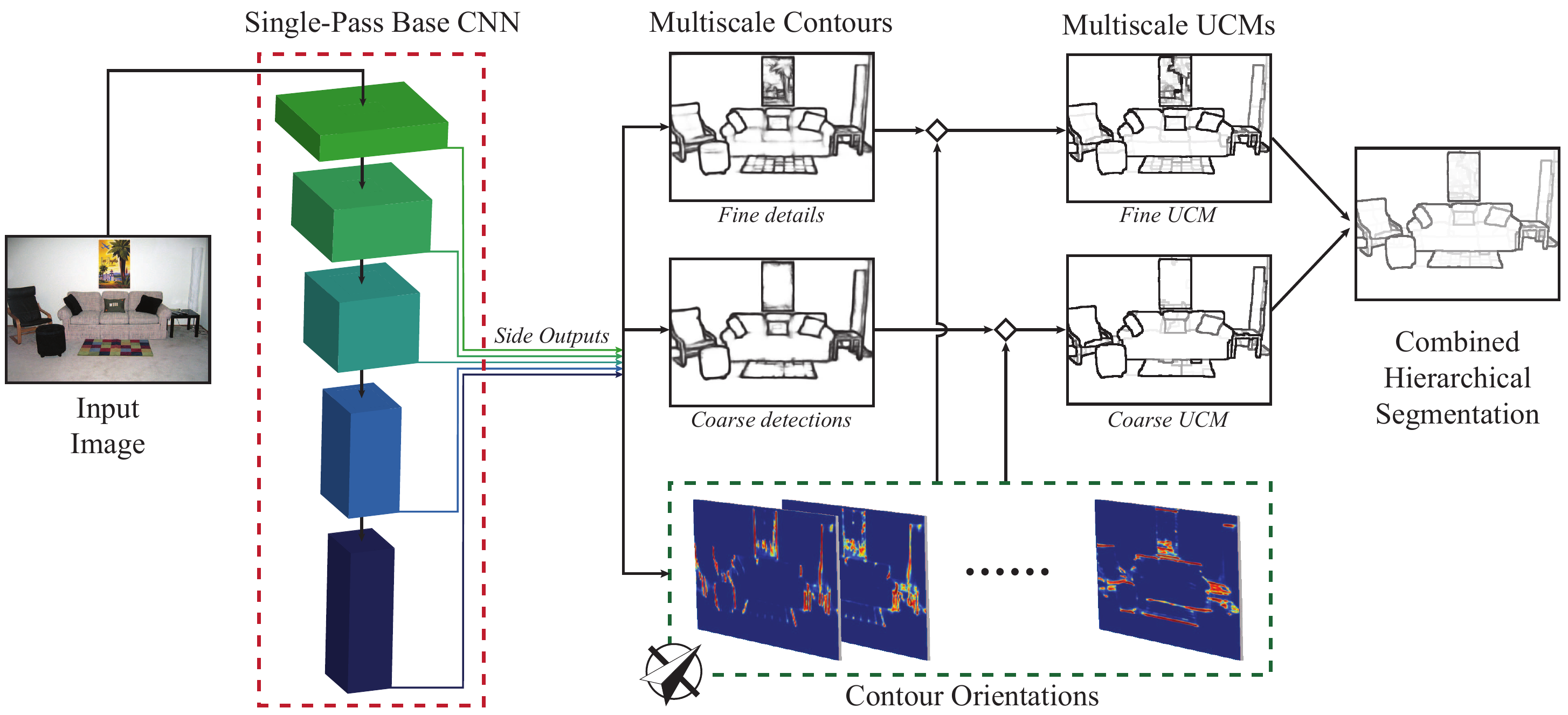}\\[-2mm]
	\caption{\textbf{Overview of COB}: From a single pass of a base CNN, we obtain multiscale oriented contours. We combine them to build Ultrametric Contour Maps (UCMs) at different scales and fuse them into a single hierarchical segmentation structure.}
\label{fig:overview}
\vspace{-4mm}
\end{figure}

The adoption of Convolutional Neural Networks (CNNs) has caused a profound change and a large leap forward in performance throughout the majority of fields in computer vision.
In the case of a traditionally category-agnostic field such as contour detection, it has recently fostered the appearance of systems~\cite{Kokkinos2016,XiTu15,BST15a,BST15b,She+15,GaLe14} that rely on large-scale category-specific information in the form of deep architectures pre-trained on Imagenet for image classification~\cite{Krizhevsky2012,Sze+15,SiZi15,He+15}.

This paper proposes Convolutional Oriented Boundaries (COB), a generic CNN architecture that allows end-to-end learning of multiscale oriented contours, and we show how it translates top performing base CNN networks into high-quality contours; allowing to bring future improvements in base CNN architectures into semantic grouping.
We then propose a sparse boundary representation for efficient construction of hierarchical regions from the contour signal. Our overall approach is both efficient (it runs in 0.8 seconds per image) and highly accurate (it produces state-of-the-art contours and regions on PASCAL and on the BSDS). Figure~\ref{fig:overview} shows an overview of our system.

For the last fifteen years, the Berkeley Segmentation Dataset and Benchmark (BSDS) \cite{Martin2001} has been the experimental testbed of choice for the study of boundary detection and image segmentation.
However, the current large-capacity and very accurate models have underlined the limitations of the BSDS as the primary benchmark for grouping. Its 300 train images are inadequate for training systems with tens of millions of parameters and, critically, current state-of-the-art techniques are reaching human performance for boundary detection on its 200 test images.

In terms of scale and difficulty, the next natural frontier for perceptual grouping is the PASCAL VOC dataset~\cite{Eve+12}, an influential benchmark for image classification, object detection, and semantic segmentation which has a \textit{trainval} set with more than 10\,000 challenging and varied images. A first step in that direction was taken by Hariharan et al.~\cite{Har+11}, who annotated the VOC dataset for category-specific boundary detection on the foreground objects. More recently, the PASCAL Context dataset~\cite{Mot+14} extended this annotation effort to all the background categories, providing thus fully-parsed images which are a direct VOC counterpart to the human ground-truth of the BSDS. 
In this direction, this paper investigates the transition from the BSDS to PASCAL Context in the evaluation of image segmentation. 

We derive valuable insights from studying perceptual grouping in a larger and more challenging empirical framework. Among them, we observe that COB leverages increasingly deeper state-of-the-art architectures, such as the recent Residual Networks~\cite{He+15}, to produce improved results. This indicates that our approach is generic and can directly benefit from future advances in CNNs. We also observe that, in PASCAL, the globalization strategy of contour strength by spectral graph partitioning proposed in~\cite{Arb+11} and used in state-of-the-art methods~\cite{Pont-Tuset2016,Kokkinos2016} is unnecessary in the presence of the high-level knowledge conveyed by pre-trained CNNs and oriented contours, thus removing a significant computational bottleneck for high-quality contours.
Overall, COB generates state-of-the-art contours and regions on PASCAL Context and on the BSDS while being computationally very efficient: it runs in 0.8 seconds per image.

We also conduct comprehensive experiments demonstrating the interest of COB for downstream recognition applications. We use our hierarchical regions as input to the combinatorial grouping algorithm of \cite{Pont-Tuset2016} and obtain state-of-the-art segmented object proposals on PASCAL Segmentation 2012 by a significant margin. Furthermore, we provide empirical evidence for the generalization power of COB by evaluating our object proposals without any retraining in the even larger and more challenging MS-COCO dataset, where we also report a large improvement in performance with respect to the state of the art.
Our efforts on segmentation through CNNs have also found application in retinal image segmentation~\cite{Man+16}, obtaining state-of-the-art and super-human performance in vessel and optic disc segmentation, which further highlights their generality.

The COB code, pre-computed results, pre-trained models, and benchmarks are publicly available at
\url{www.vision.ee.ethz.ch/~cvlsegmentation/}.

\section{Related Work}
\label{sec:related}
The latest wave of contour detectors takes advantage of deep learning to obtain state-of-the-art results~\cite{Kokkinos2016,XiTu15,BST15a,BST15b,She+15,GaLe14,BST15c}.
Ganin and Lempitsky~\cite{GaLe14} use a deep architecture to extract features of image patches. They approach contour detection as a multiclass classification task, by matching the extracted features to predefined ground-truth features. The authors of~\cite{BST15a,BST15b} make use of features generated by pre-trained CNNs to regress contours. They prove that object-level information provides powerful cues for the prediction of contours. Shen et al.~\cite{She+15} learn deep features using shape information. 
Xie and Tu~\cite{XiTu15} provide an end-to-end deep framework to boost the efficiency and accuracy of contour detection, using convolutional feature maps and a novel loss function.
Kokkinos~\cite{Kokkinos2016} builds upon \cite{XiTu15} and improves the results by tuning the loss function, running the detector at multiple scales, and adding globalization.
COB is different from this previous work in that we obtain multiscale information in a single pass of the network on the whole image, it combines the per-pixel classification with contour orientation estimation, and its output is richer than a linear combination of cues at different scales.

At the core of all these deep learning approaches, lies a \textit{base CNN}, starting from the seminal AlexNet~\cite{Krizhevsky2012} (8 layers), through the more complex VGGNet~\cite{SiZi15} (16 layers) and inception architecture of GoogLeNet~\cite{Sze+15} (22 layers), to the very recent and very deep ResNets~\cite{He+15} (up to 1001 layers).
Image classification results, which originally motivated these architectures, have been continuously improved by exploring deeper and more complex networks. In this work, we present results both using VGGNet and ResNet, showing that COB is modular and can incorporate and benefit from future improvements in the base CNN.

Recent work has also explored the weakly supervised or unsupervised learning of contours: Khoreva et al.~\cite{Kho+16} learn from the results of generic contour detectors coupled with object detectors; and Li et al.~\cite{Li+16} train contour detectors from motion boundaries acquired from video sequences.
Yang et al.~\cite{Yan+16} use conditional random fields to refine the inaccurately localized boundary annotations of PASCAL.
Our approach uses full supervision from BSDS and PASCAL Context for contour localization and orientation.

COB exploits the duality between contour detection and segmentation hierarchies, initially studied by Najman and Schmitt~\cite{Najman1996}. Arbel\'aez et al.~\cite{Arb+11} showed its usefulness for jointly optimizing contours and regions. Pont-Tuset et al.~\cite{Pont-Tuset2016} leveraged multi-resolution contour detection and proved its interest also for generating object proposals.
We differentiate from these approaches in two aspects. First, our sparse boundary representation translates into a clean and highly efficient implementation of hierarchical segmentation. Second, by leveraging high-level knowledge from the CNNs in the estimation of contour strength and orientation, our method benefits naturally from global information, which allows bypassing the globalization step (output of normalized cuts), a bottleneck in terms of computational cost, but a cornerstone of previous aproaches.

\section{Deep Multiscale Oriented Contours}
\label{sec:hier_cont}

CNNs are by construction multi-scale feature extractors.  If one examines the standard architecture of a CNN consisting of convolutional and spatial pooling layers, it becomes clear that as we move deeper, feature maps capture more global information due to the decrease in resolution. For contour detection, this architecture implies local and fine-scale contours at shallow levels, coarser spatial resolution and larger receptive fields for the units when going deeper into the network and, consequently, more global information for predicting boundary strength and orientation. CNNs have therefore a built-in globalization strategy for contour detection, analogous to the hand-engineered globalization of contour strength through spectral graph partitioning in~\cite{Arb+11,Pont-Tuset2016}.

\begin{figure}[b]
\includegraphics[width=\linewidth]{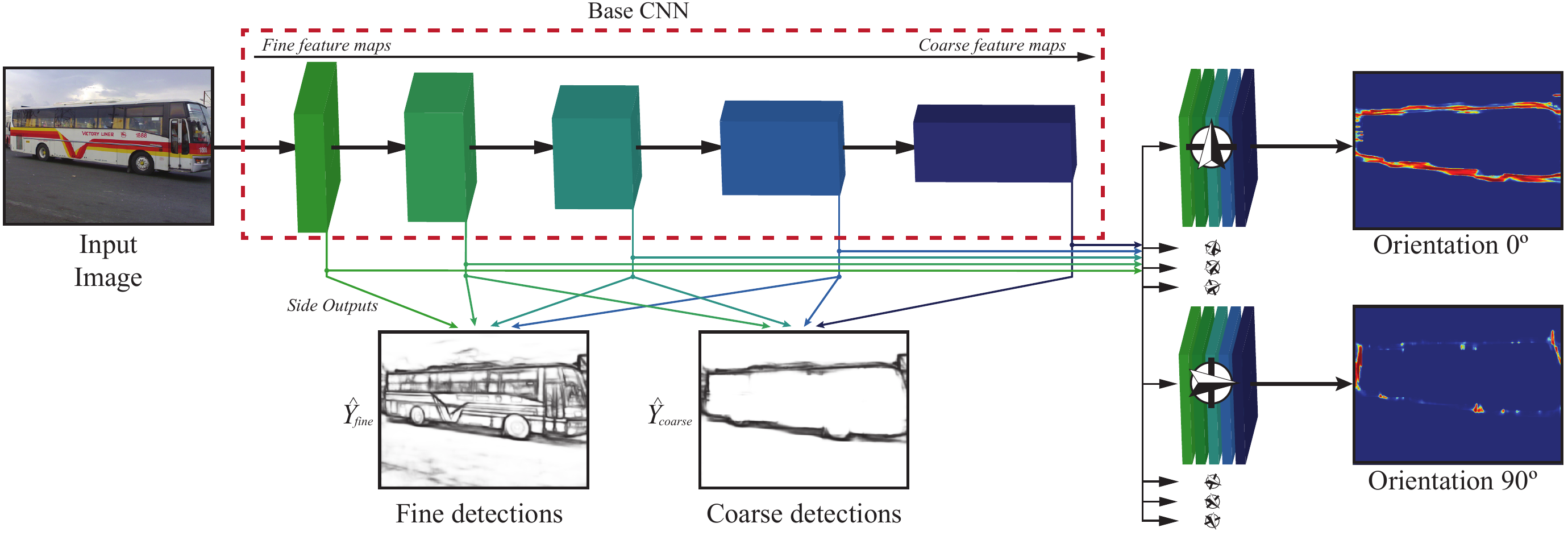}\\[-8mm]
\caption{\textbf{Our deep learning architecture} (best viewed in color). The connections show the different stages that are used to generate the multiscale contours. Orientations further require additional convolutional layers in multiple stages of the network.}
\label{fig:CNN}
\end{figure}

Figure~\ref{fig:CNN} depicts how we make use of information provided by the intermediate layers of a CNN to detect contours and their orientations at multiple scales. Different groups of feature maps contain different, scale-specific information, which we combine to build a multiscale oriented contour detector.
The remainder of this section is devoted to introducing the recent approaches to contour detection using deep learning, to presenting our CNN architecture to produce contour detection at different scales, and to explain how we estimate the orientation of the edges; all in a single CNN forward pass at the image level.

\paragraph*{\textbf{Training deep contour detectors:}}
The recent success of~\cite{XiTu15} is based on a CNN to accurately regress the contours of an image. Within this framework, the idea of employing a neural network in an image-to-image fashion without any post-processing has proven successful and serves right now as the state-of-the-art for the task of contour detection. Their network, HED, produces scale-specific contour images (side outputs) for different scales of a network, and combines their activations linearly to produce a contour probability map. Using the notation of the authors, we denote the training dataset by $S=\lbrace \left( X_n, Y_n \right),n=1,\dots,N \rbrace$, with $X_n$ being the input image and $Y_n=\lbrace y_j^{(n)},j=1,\dots,|X_{n}| \rbrace ,  y_j^{(n)} \in \lbrace 0,1 \rbrace$  the predicted pixelwise labels. For simplicity, we drop the subscript $n$. Each of the $M$ side outputs minimizes the objective function:
\footnotesize
\begin{align}
\ell_{side}^{(\!m\!)\!}\!\!\left( \!\mathbf{W}\!,\!\mathbf{w}^{\!(\!m\!)\!}\!\right)\!=\!-\!\beta\!\!\sum_{j	\in Y_+}\!{\!\log\!{P\!\left(\!y_j\!=\!1 |X;\!\mathbf{W}\!,\!\mathbf{w}^{\!(\!m\!)\!}\!\right)}}\!-\!(1\!-\!\beta)\!\!\sum_{j	\in Y_-}\!{\!\log{\!P\!\left(\!y_j\!=\!0 |X;\!\mathbf{W}\!,\!\mathbf{w}^{\!(\!m\!)\!}\!\right)}} \label{eq:hed_cost_side}
\end{align}
\normalsize
where $\ell_{side}^{(m)}$ is the loss function for scale $m \in \lbrace 1,\dots,M \rbrace$, $\mathbf{W}$ denotes the standard set of parameters of the CNN, and $\lbrace \mathbf{w}^{(m)}, m=1,\dots,M \rbrace$ the corresponding weights of the the $m$-th side output. The multiplier $\beta$ is used to handle the imbalance of the substantially greater number of background compared to contour pixels. $Y_+$ and $Y_-$ denote the contour and background sets of the ground-truth $Y$, respectively.  The probability $P \left( \cdot \right)$ is obtained by applying a sigmoid $\sigma \left( \cdot \right)$ to the activations of the side outputs $\hat{A}_{side}^{(m)}=\lbrace a_j^{(m)}, j=1,\dots,|Y| \rbrace$. The activations are finally fused linearly, as: $\hat{Y}_{fuse} = \sigma \left( \Sigma_{m=1}^{M}h_m \hat{A}_{side}^{(m)} \right)$ where $\mathbf{h}=\lbrace h_m,m=1,\dots,M \rbrace$ are the fusion weights. The fusion output is also trained to resemble the ground-truth applying the same loss function of Equation~\ref{eq:hed_cost_side}, by optimizing the complete set of parameters, including the fusion weights $\mathbf{h}$. 
In the rest of the paper we use the class-balancing cross-entropy loss function of Equation~\ref{eq:hed_cost_side}.

\paragraph*{\textbf{Multiscale contours:}}

We finetune the 50-layer ResNet~\cite{He+15} for the task of contour detection. The fully connected layers used for classification are removed, and so are the batch normalization layers, since we operate on one image per iteration. Therefore, the network consists mainly of convolutional layers coupled with ReLU activations, divided into 5 stages. We will refer to this architecture as the ``base CNN'' of our implementation. Each stage is handled as a different scale, since it contains feature maps of a similar size. At the end of a stage, there is a max pooling layer, which reduces the dimensions of the produced feature maps to a half. As discussed before, the CNN naturally contains multiscale information, which we exploit to build a multiscale contour regressor.

We separately supervise the output of the last layer of each stage (side activation), comparing it to the ground truth using the loss function of Equation~\ref{eq:hed_cost_side}.
This way, we enforce each side activation to produce an intermediate contour map at different resolution.
The idea of supervising intermediate parts of a CNN has successfully been used in previous approaches, for a variety of tasks~\cite{Sze+15,Lee+14,XiTu15}.
In the 5-scale base CNN illustrated in Figure~\ref{fig:CNN}, we linearly combine the side activations 
of the 4 finest and 4 coarsest scales to a fine-scale and a coarse-scale output ($\hat{Y}_{fine}$ and $\hat{Y}_{coarse}$, respectively) with trainable weights.
The finer scale contains better localized contours, whereas the coarse scale leads to less noisy detections.
To train the two sets of weights of the linear combinations, we freeze the pre-trained weights of the base CNN.

\paragraph*{\textbf{Estimation of Contour Orientations:}}
In order to predict accurate contour orientations, we propose an extension of the CNN that we use as multiscale contour detector. We define the task as pixel-wise image-to-image multiscale classification into $K$ bins. We connect $K$ different branches (sub-networks) to the base network, each of which is associated with one orientation bin, and has access to feature maps that are generated from the intermediate convolutional layers at $M$ different scales. We assign the parts of the CNN associated with each orientation a different task than the base network: 
classify the pixels of the contours that match a specific orientation. In order to design these orientation-specific subtasks, we classify each pixel of the human contour annotations into $K$ different orientations. The orientation of each contour pixel is obtained by approximating the ground-truth boundaries with polygons, and assigning each pixel the orientation of the closest polygonal segment, as shown in Figure~\ref{fig:pol_simpl}. As in the case of multiscale contours, the weights of the base network remain frozen when training these sub-networks.

\begin{figure}[h!]
\setlength{\fboxsep}{0pt}
\centering
\cfbox{white}{\resizebox{0.23\textwidth}{!}{%
\begin{tikzpicture}
    \node[anchor=north west,inner sep=0,use as bounding box] (image) at (0,0) {\includegraphics[width=\textwidth]{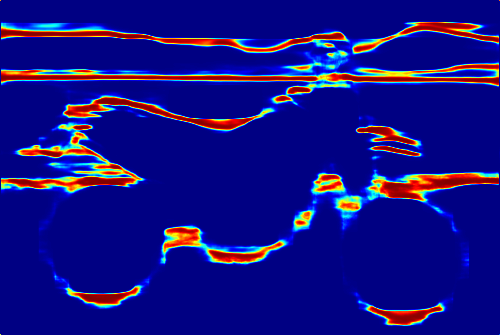}}; 
        \clip(image.north west) rectangle (image.south east);
     \node[rotate around={0:(0,0)}] at (1.2,-1.2) {\includegraphics[width=.2\textwidth]{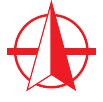}};
\end{tikzpicture}}}
\cfbox{white}{\resizebox{0.23\textwidth}{!}{%
\begin{tikzpicture}
    \node[anchor=north west,inner sep=0,use as bounding box] (image) at (0,0) {\includegraphics[width=\textwidth]{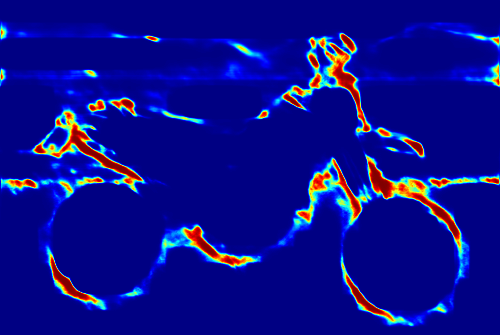}}; 
        \clip(image.north west) rectangle (image.south east);
     \node[rotate around={-45:(0,0)}] at (1.2,-1.2) {\includegraphics[width=.2\textwidth]{img/arrow.pdf}};
\end{tikzpicture}}}
\cfbox{white}{\resizebox{0.23\textwidth}{!}{%
\begin{tikzpicture}
    \node[anchor=north west,inner sep=0,use as bounding box] (image) at (0,0) {\includegraphics[width=\textwidth]{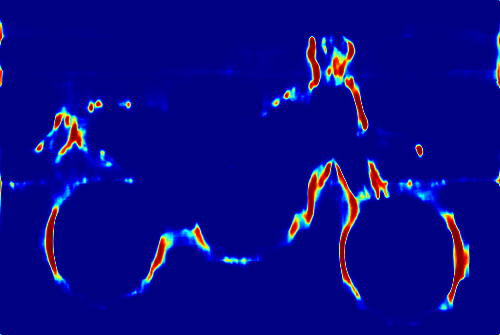}};
        \clip(image.north west) rectangle (image.south east);
     \node[rotate around={-90:(0,0)}] at (1.2,-1.2) {\includegraphics[width=.2\textwidth]{img/arrow.pdf}};
\end{tikzpicture}}}
\cfbox{white}{\resizebox{0.23\textwidth}{!}{%
\begin{tikzpicture}
    \node[anchor=north west,inner sep=0,use as bounding box] (image) at (0,0) {\includegraphics[width=\textwidth]{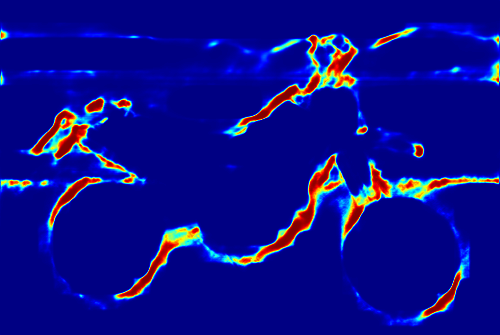}}; 
    \clip(image.north west) rectangle (image.south east);
     \node[rotate around={-135:(0,0)}] at (1.2,-1.2) {\includegraphics[width=.2\textwidth]{img/arrow.pdf}};
\end{tikzpicture}}}
\\[1mm]
\fbox{\resizebox{0.23\textwidth}{!}{%
\begin{tikzpicture}
    \node[anchor=south west,inner sep=0] (image) at (0,0) {\includegraphics[width=\textwidth]{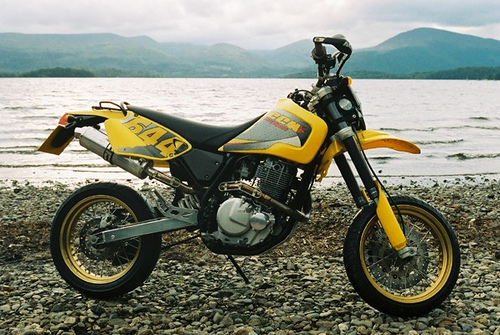}};    
\end{tikzpicture}}}
\fbox{\resizebox{0.23\textwidth}{!}{%
\begin{tikzpicture}
    \node[anchor=south west,inner sep=0] (image) at (0,0) {\includegraphics[width=\textwidth]{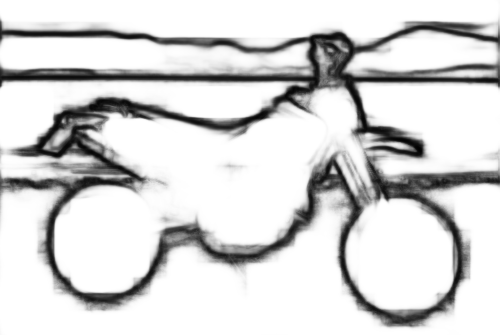}};  
\end{tikzpicture}}}
\fbox{\resizebox{0.23\textwidth}{!}{%
\begin{tikzpicture}
    \node[anchor=south west,inner sep=0] (image) at (0,0) {\includegraphics[width=0.95\textwidth]{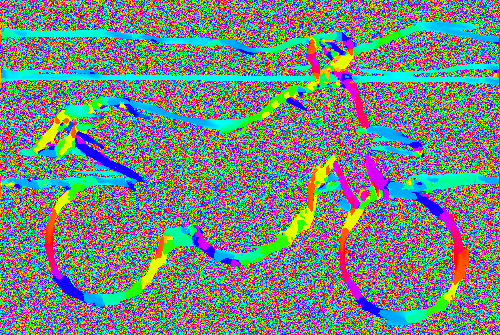}};  
\end{tikzpicture}}}
\fbox{\resizebox{0.23\textwidth}{!}{%
\begin{tikzpicture}
    \node[anchor=south west,inner sep=0] (image) at (0,0) {\includegraphics[width=\textwidth]{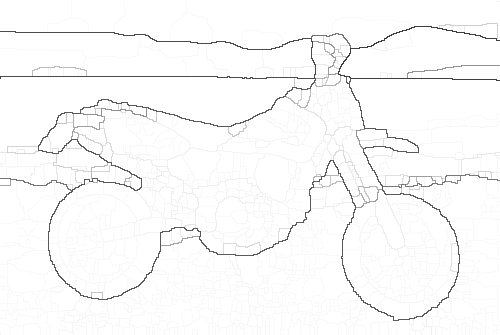}}; 
\end{tikzpicture}}} \\
	\caption{Illustration of contour orientation learning. Row 1 shows the responses $B_k$ for 4 out of the 8 orientation bins. Row 2, from left to right: original image, contour strength, learned orientation map into 8 orientations, and hierarchical boundaries.}
\label{fig:orientations}
\end{figure}

Each sub-network consists of $M$ convolutional layers, each of them appended on different scales of the base network. Thus we need $M*K$ additional layers, namely \texttt{conv\_scale\_m\_orient\_k}, with \texttt{k}$=1,\dots,K$ and \texttt{m}$=1,\dots,M$. In our setup, we use $K=8$ and $M=5$. All $K$ orientations are regressed in parallel, and since they are associated with a certain angle, we post-process them to obtain the orientation map. Specifically, the orientation map is obtained as:
\begin{equation}
O(x,y) = \mathcal{T}\left(\arg\max_k{B_k \left(x,y \right)}\right),  k=1,\dots,K
\end{equation}
where $B_k(x,y)$ denotes the response of the $k$-th orientation bin of the CNN at the pixels with coordinates  $(x,y)$  and $\mathcal{T}\left(\cdot\right)$ is the transformation function which associates each bin with its central angle. For the cases where two neighboring bins lead to strong responses, we compute the angle as their weighted average. At pixels where there is no response for any of the orientations, we assign random values between $0$ and $\pi$, not to bias the orientations. The different orientations as well as the resulting orientation map (color-coded) are illustrated in Figure~\ref{fig:orientations}. 

In \cite{Arb+11,DoZi13,Pont-Tuset2016} the orientations are computed by means of local gradient filters.  In Section~\ref{sec:experiments} we show that our learned orientations are significantly more accurate and lead to more better region segmentations.

\section{Fast Hierarchical Regions}
\label{sec:fast_mcg}
This section is devoted to building an efficient hierarchical image segmentation algorithm from the multiscale contours and the orientations extracted in the previous section.
We build on the concept of Ultrametric Contour Map (UCM)~\cite{Arb+11}, which transforms a contour detection probability map into a hierarchical boundary map, which gets partitions at different granularities when thresholding at various contour strength values.
Despite the success of UCMs, their low speed significantly limits their applicability.

In the remainder of this section we first describe an alternative representation of an image partition that allows us to reduce the computation time of multiscale UCMs by an order of magnitude, to less than one second.
Then, we present the global algorithm to build a hierarchy of regions from the
multiscale contours and the orientations presented in Section~\ref{sec:hier_cont}.
As we will show in the experimental section, the resulting algorithm improves the state of the art significantly, at a fraction of the computational time of~\cite{Pont-Tuset2016}. 

\begin{figure}
\centering
\begin{minipage}{0.68\linewidth}
\resizebox{\textwidth}{!}{\includegraphics{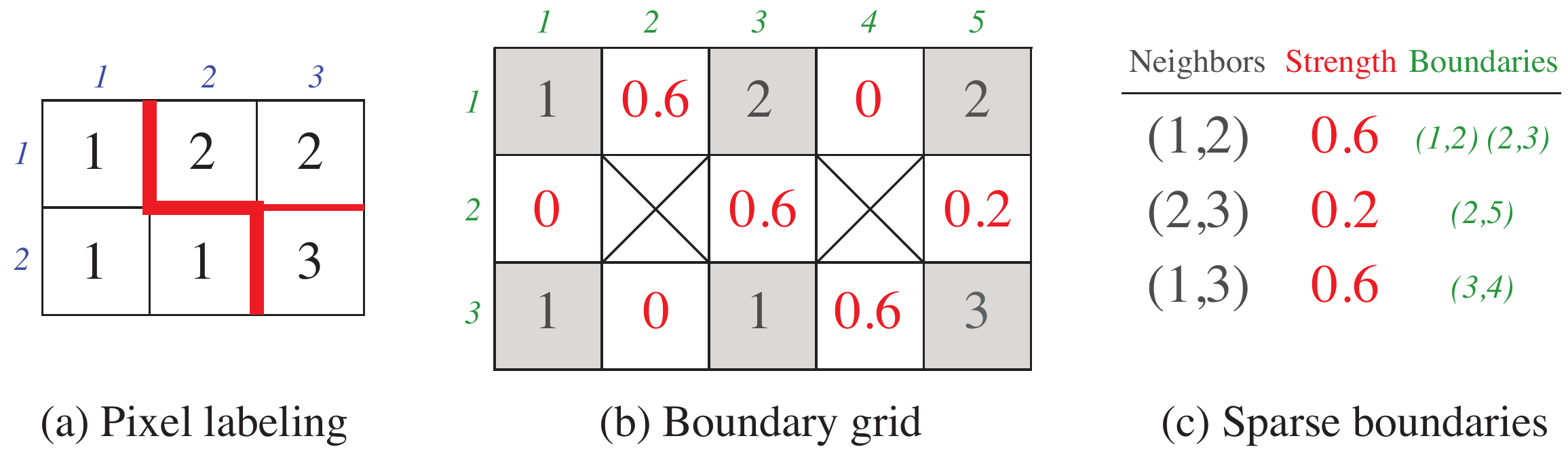}}
\vspace{-5mm}
\caption{\textbf{Image Partition Representation:}\newline(a) Pixel labeling, each pixel gets assigned a region label. (b) Boundary grid, markers of the boundary positions.\newline(c) Sparse boundaries, lists of boundary coordinates between neighboring regions.}
\label{fig:part_rep}
\end{minipage}
\hfill
\begin{minipage}{0.28\linewidth}
\centering
\resizebox{\textwidth}{!}{\includegraphics{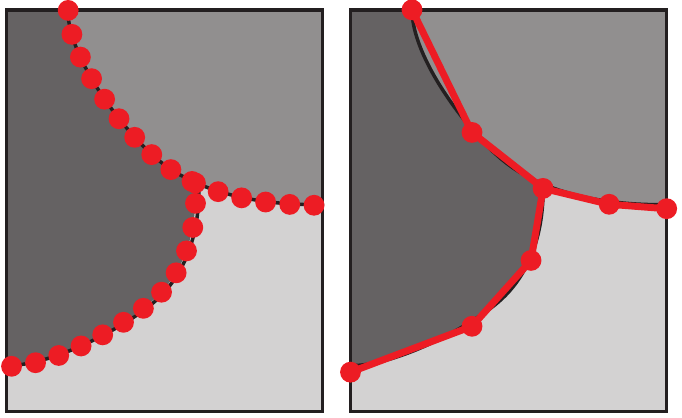}}
\vspace{-3mm}
\caption{\textbf{Polygon simplification:} From all boundary points (left) to simplified polygons (right).}
\label{fig:pol_simpl}
\end{minipage}
\end{figure}

\paragraph*{\textbf{Sparse Boundary Representation of Hierarchies of Regions:}}
An image partition is a clustering of the set of pixels into different sets,
which we call regions.
The most straightforward way of representing it in a computer is by a matrix of labels, as in the
example in Figure~\ref{fig:part_rep}(a), with three regions on an image of size 2$\times$3.
The boundaries of this partition are the edge elements, or \textit{edgels}, between the pixels with different
labels (highlighted in red).
We can assign different \textit{strengths} to these boundaries (thicknesses of the red lines), which indicate the \textit{confidence} of that piece of being a true boundary.
By iteratively \textit{erasing} these boundaries in order of increasing strength
we obtain different partitions, which we call \textit{hierarchy of regions}, or Ultrametric Contour Maps.

\label{sec:cont_ucm}

These boundaries are usually stored in the \textit{boundary grid} (Figure~\ref{fig:part_rep}(b)), a matrix
of double the size of the image (minus one), in which the odd coordinates represent pixels (gray areas), and the positions in between represent boundaries (red numbers) and junctions (crossed positions).

UCMs use this representation to store their boundary \textit{strength} values,
that is, each boundary position stores the threshold value beyond which that edgel \textit{disappears} and the two neighboring regions merge.
This way, simply \textit{binarizing} a UCM we have a partition represented as a boundary grid.

This representation, while useful during prototyping, becomes very inefficient at run time, where the percentage of \textit{activated} boundaries is very sparse.
Not only are we wasting memory by storing those \textit{empty} boundaries, but it also makes 
operating on them very inefficient by having to \textit{sweep} over the entire matrix 
to perform a modification on a single boundary piece.

Inspired by how sparse matrices are handled, we designed the \textit{sparse boundaries} representation
(Figure~\ref{fig:part_rep}(c)).
It stores a look-up table for pairs of neigboring regions, their boundary strength, and the list of coordinates the boundary occupies.
Apart from being more compact in terms of memory, this representation enables efficient operations on specific pieces of a boundary, since one only needs to perform a
search in the look-up table and scan the activated coordinates; instead of having to sweep the whole boundary grid.

\paragraph*{\textbf{Fast Hierarchies from Multiscale Oriented Contours:}}
The deep CNN presented in Section~\ref{sec:hier_cont} provides different levels of detail for the image contours. 
A linear combination of the layers is the straightforward way of providing a single
contour signal~\cite{XiTu15}.
The approach in this work is to combine the region hierarchies extracted from 
the contour signals at each layer instead of the contours directly.
We were inspired by the framework proposed in~\cite{Pont-Tuset2016},
in which a UCM is obtained from contours computed at different image scales and then combined in a single hierarchy; but instead we use the different contour outputs that are computed in a single pass of the proposed CNN architecture.

A drawback of the original framework~\cite{Pont-Tuset2016} is that the manipulation of the hierarchies is very slow (in the order of seconds), so the operations on the UCMs had to be discretized and performed at a low number of contour strengths.
By using the fast sparse boundary representation, we can operate on all contour strengths, yielding better results at a fraction of the original cost. 
Moreover, we use the learned contour orientations for the computation of the Oriented Watershed Transform (OWT), further boosting performance.

\section{Experiments}
\label{sec:experiments}
This section presents the empirical evidence that supports our approach.
First, Section~\ref{sec:exp:ablat} explores the ablated and baseline techniques studied to isolate and quantify the improvements due to different components of our system.
Then Section~\ref{sec:exp:orient}, Section~\ref{sec:exp:generic}, and Section~\ref{sec:exp:prop} compare our results against the state-of-the-art in terms of contour orientation estimation, generic image segmentation, and the application to object proposals, respectively.
In all three cases, we obtain the best results to date by a significant margin.
Finally, Section~\ref{sec:exp:speed} analyzes the gain in speed achieved mainly by the use of our sparse boundaries representation.

We extend the main BSDS benchmarks to the PASCAL Context dataset~\cite{Mot+14}, which contains carefully localized pixelwise semantic annotations for the entire image on the PASCAL VOC 2010 detection trainval set. 
This results in 459 semantic categories across 10\,103 images, which is an order of magnitude (20$\times$) larger than the BSDS. In order to allow training and optimization of large capacity models, we split the data into train, validation, and test sets as follows: \emph{VOC train} corresponds to the official PASCAL Context train with 4\,998 images, \emph{VOC val} corresponds to half the official PASCAL Context validation set with 2\,607 images and \emph{VOC test} corresponds to the second half with 2\,498 images. In the remainder of the paper,
we refer to this dataset division. Note that, in this setting, the notion of boundary is defined as separation between different semantic categories and not their parts, in contrast to the BSDS.

We used the publicly available \textit{Caffe}~\cite{Jia+14} framework for training and testing CNNs, and all the state-of-the-art results are computed using the publicly-available code provided by the respective authors.

\subsection{Control Experiments/Ablation Analysis}
\label{sec:exp:ablat}
This section presents the control experiments and ablation analysis to assess the performance of all subsystems of our method.
We train on \emph{VOC train}, and evaluate on \emph{VOC val} set.
We report the standard F-measure at Optimal Dataset Scale (ODS) and Optimal Image Scale (OIS), as well as the Average Precision (AP), both evaluating boundaries ($F_b$~\cite{Martin2004}) and regions ($F_\mathit{op}$~\cite{Pont-Tuset2016a}).

Table~\ref{table:ablation} shows the evaluation results of the different variants, 
highlighting whether we include globalization and/or trained orientations. 
As a first baseline, we test the performance of MCG~\cite{Pont-Tuset2016}, which uses Structured Edges~\cite{DoZi13} as input contour signal, and denote it MCG~\cite{Pont-Tuset2016}.
We then substitute SE by the newer HED~\cite{XiTu15}, trained on \emph{VOC train} as input contours and denote it MCG-HED.
Note that the aforementioned baselines require multiple passes of the contour detector (3 different scales).

In the direction of using the side outputs of the base CNN architecture as multiscale contour detections in one pass, we tested the baseline of naively taking the 5 side outputs 
directly as the contour detections. We trained both VGGNet~\cite{SiZi15} and ResNet50~\cite{He+15} on \emph{VOC train} and combined the 5 side outputs with 
our fast hierarchical regions of Section~\ref{sec:fast_mcg} (VGGNet-Side and ResNet50-Side).

We finally evaluate different variants of our system, as presented in Section~\ref{sec:hier_cont}. 
We first compare our system with two different base architectures: Ours(VGGNet) and Ours(ResNet50). We train the base networks for 30000 iterations, with stochastic gradient descent and a momentum of 0.9. We observe that the deeper architecture of ResNet translates into better boundaries and regions.

We then evaluate the influence of our trained orientations and globalization, by testing 
the four possible combinations (the orientations are further evaluated in next section).
Our method using ResNet50 together with trained orientations leads to the best results both for boundaries and for regions.
The experiments also show that, when coupled with trained orientations, globalization even decreases performance, so we can safely remove it and get a significant speed up.
Our technique with trained orientations and without globalization is therefore selected as our final system and will be referred to in the sequel as Convolutional Oriented Boundaries (COB).

\begin{figure}[t]
\begin{minipage}[b]{0.57\linewidth}
\setlength{\tabcolsep}{4pt} 
\center
\footnotesize
\resizebox{\textwidth}{!}{%
\begin{tabular}{lcccccccc}
\toprule
       & & &\multicolumn{3}{c}{Boundaries - $F_b$} & \multicolumn{3}{c}{Regions - $F_\mathit{op}$}\\
Method & Global. & Orient. & ODS & OIS & AP & ODS & OIS & AP \\
\midrule
MCG~\cite{Pont-Tuset2016}   & {\color{gray}\ding{51}} &  {\color{gray}\ding{55}} & %
\mbox{\input{data/pr/PASCALContext_val_new_fb_MCG-BSDS500_ods_f.txt}\hspace{-2.5pt}}%
 & %
\mbox{\input{data/pr/PASCALContext_val_new_fb_MCG-BSDS500_ois_f.txt}\hspace{-2.5pt}}%
 & %
\mbox{\input{data/pr/PASCALContext_val_new_fb_MCG-BSDS500_ap.txt}\hspace{-2.5pt}}%
 & %
\mbox{\input{data/pr/PASCALContext_val_new_fop_MCG-BSDS500_ods_f.txt}\hspace{-2.5pt}}%
 & %
\mbox{\input{data/pr/PASCALContext_val_new_fop_MCG-BSDS500_ois_f.txt}\hspace{-2.5pt}}%
 & %
\mbox{\input{data/pr/PASCALContext_val_new_fop_MCG-BSDS500_ap.txt}\hspace{-2.5pt}}%
 \\
MCG-HED  & {\color{gray}\ding{51}} &  {\color{gray}\ding{55}} & %
\mbox{\input{data/pr/PASCALContext_val_new_fb_MCG-HED_ods_f.txt}\hspace{-2.5pt}}%
 & %
\mbox{\input{data/pr/PASCALContext_val_new_fb_MCG-HED_ois_f.txt}\hspace{-2.5pt}}%
 & %
\mbox{\input{data/pr/PASCALContext_val_new_fb_MCG-HED_ap.txt}\hspace{-2.5pt}}%
 & %
\mbox{\input{data/pr/PASCALContext_val_new_fop_MCG-HED_ods_f.txt}\hspace{-2.5pt}}%
 & %
\mbox{\input{data/pr/PASCALContext_val_new_fop_MCG-HED_ois_f.txt}\hspace{-2.5pt}}%
 & %
\mbox{\input{data/pr/PASCALContext_val_new_fop_MCG-HED_ap.txt}\hspace{-2.5pt}}%
\\
\midrule
VGGNet-Side & {\color{gray}\ding{51}} & {\color{gray}\ding{55}} & %
\mbox{\input{data/pr/PASCALContext_val_new_fb_VGGNet-Side_ods_f.txt}\hspace{-2.5pt}}%
 & %
\mbox{\input{data/pr/PASCALContext_val_new_fb_VGGNet-Side_ois_f.txt}\hspace{-2.5pt}}%
 & %
\mbox{\input{data/pr/PASCALContext_val_new_fb_VGGNet-Side_ap.txt}\hspace{-2.5pt}}%
 & %
\mbox{\input{data/pr/PASCALContext_val_new_fop_VGGNet-Side_ods_f.txt}\hspace{-2.5pt}}%
 & %
\mbox{\input{data/pr/PASCALContext_val_new_fop_VGGNet-Side_ois_f.txt}\hspace{-2.5pt}}%
 & %
\mbox{\input{data/pr/PASCALContext_val_new_fop_VGGNet-Side_ap.txt}\hspace{-2.5pt}}%
\\
ResNet50-Side & {\color{gray}\ding{51}} & {\color{gray}\ding{55}} & %
\mbox{\input{data/pr/PASCALContext_val_new_fb_ResNet50-Side_ods_f.txt}\hspace{-2.5pt}}%
 & %
\mbox{\input{data/pr/PASCALContext_val_new_fb_ResNet50-Side_ois_f.txt}\hspace{-2.5pt}}%
 & %
\mbox{\input{data/pr/PASCALContext_val_new_fb_ResNet50-Side_ap.txt}\hspace{-2.5pt}}%
 & %
\mbox{\input{data/pr/PASCALContext_val_new_fop_ResNet50-Side_ods_f.txt}\hspace{-2.5pt}}%
 & %
\mbox{\input{data/pr/PASCALContext_val_new_fop_ResNet50-Side_ois_f.txt}\hspace{-2.5pt}}%
 & %
\mbox{\input{data/pr/PASCALContext_val_new_fop_ResNet50-Side_ap.txt}\hspace{-2.5pt}}%
\\
\midrule
Ours (VGGNet) & \ding{55} & \ding{51} & %
\mbox{\input{data/pr/PASCALContext_val_new_fb_COB_train-HED_ods_f.txt}\hspace{-2.5pt}}%
 & %
\mbox{\input{data/pr/PASCALContext_val_new_fb_COB_train-HED_ois_f.txt}\hspace{-2.5pt}}%
 & %
\mbox{\input{data/pr/PASCALContext_val_new_fb_COB_train-HED_ap.txt}\hspace{-2.5pt}}%
 & %
\mbox{\input{data/pr/PASCALContext_val_new_fop_COB_train-HED_ods_f.txt}\hspace{-2.5pt}}%
 & %
\mbox{\input{data/pr/PASCALContext_val_new_fop_COB_train-HED_ois_f.txt}\hspace{-2.5pt}}%
 & %
\mbox{\input{data/pr/PASCALContext_val_new_fop_COB_train-HED_ap.txt}\hspace{-2.5pt}}%
\\
\midrule
Ours (ResNet50) & \ding{55} & \ding{55} & %
\mbox{\input{data/pr/PASCALContext_val_new_fb_COB_train-noglob-noorient_ods_f.txt}\hspace{-2.5pt}}%
 & %
\mbox{\input{data/pr/PASCALContext_val_new_fb_COB_train-noglob-noorient_ois_f.txt}\hspace{-2.5pt}}%
 & %
\mbox{\input{data/pr/PASCALContext_val_new_fb_COB_train-noglob-noorient_ap.txt}\hspace{-2.5pt}}%
 & %
\mbox{\input{data/pr/PASCALContext_val_new_fop_COB_train-noglob-noorient_ods_f.txt}\hspace{-2.5pt}}%
 & %
\mbox{\input{data/pr/PASCALContext_val_new_fop_COB_train-noglob-noorient_ois_f.txt}\hspace{-2.5pt}}%
 & %
\mbox{\input{data/pr/PASCALContext_val_new_fop_COB_train-noglob-noorient_ap.txt}\hspace{-2.5pt}}%
\\
Ours (ResNet50) & \ding{51} & \ding{55} & %
\mbox{\input{data/pr/PASCALContext_val_new_fb_COB_train-glob-noorient_ods_f.txt}\hspace{-2.5pt}}%
 & %
\mbox{\input{data/pr/PASCALContext_val_new_fb_COB_train-glob-noorient_ois_f.txt}\hspace{-2.5pt}}%
 & %
\mbox{\input{data/pr/PASCALContext_val_new_fb_COB_train-glob-noorient_ap.txt}\hspace{-2.5pt}}%
 & %
\mbox{\input{data/pr/PASCALContext_val_new_fop_COB_train-glob-noorient_ods_f.txt}\hspace{-2.5pt}}%
 & %
\mbox{\input{data/pr/PASCALContext_val_new_fop_COB_train-glob-noorient_ois_f.txt}\hspace{-2.5pt}}%
 & %
\mbox{\input{data/pr/PASCALContext_val_new_fop_COB_train-glob-noorient_ap.txt}\hspace{-2.5pt}}%
\\
Ours (ResNet50) & \ding{51} & \ding{51} & %
\mbox{\input{data/pr/PASCALContext_val_new_fb_COB_train-glob-orient_ods_f.txt}\hspace{-2.5pt}}%
 & %
\mbox{\input{data/pr/PASCALContext_val_new_fb_COB_train-glob-orient_ois_f.txt}\hspace{-2.5pt}}%
 & %
\mbox{\input{data/pr/PASCALContext_val_new_fb_COB_train-glob-orient_ap.txt}\hspace{-2.5pt}}%
 & %
\mbox{\input{data/pr/PASCALContext_val_new_fop_COB_train-glob-orient_ods_f.txt}\hspace{-2.5pt}}%
 & \bf%
\mbox{\input{data/pr/PASCALContext_val_new_fop_COB_train-glob-orient_ois_f.txt}\hspace{-2.5pt}}%
 & \bf%
\mbox{\input{data/pr/PASCALContext_val_new_fop_COB_train-glob-orient_ap.txt}\hspace{-2.5pt}}%
\\
Ours (ResNet50) & \ding{55} & \ding{51} & \bf%
\mbox{\input{data/pr/PASCALContext_val_new_fb_COB_train-noglob-orient_ods_f.txt}\hspace{-2.5pt}}%
 & \bf%
\mbox{\input{data/pr/PASCALContext_val_new_fb_COB_train-noglob-orient_ois_f.txt}\hspace{-2.5pt}}%
 & \bf%
\mbox{\input{data/pr/PASCALContext_val_new_fb_COB_train-noglob-orient_ap.txt}\hspace{-2.5pt}}%
 & \bf%
\mbox{\input{data/pr/PASCALContext_val_new_fop_COB_train-noglob-orient_ods_f.txt}\hspace{-2.5pt}}%
 & %
\mbox{\input{data/pr/PASCALContext_val_new_fop_COB_train-noglob-orient_ois_f.txt}\hspace{-2.5pt}}%
 & %
\mbox{\input{data/pr/PASCALContext_val_new_fop_COB_train-noglob-orient_ap.txt}\hspace{-2.5pt}}%
\\
\bottomrule
\end{tabular}}
\vspace{2mm}
\captionof{table}{\textbf{Ablation analysis on \emph{VOC val}}: Comparison of different ablated versions of our system.}
\label{table:ablation}
\end{minipage}
\hfill
\begin{minipage}[b]{0.38\linewidth}
\centering
\hspace{-2mm}
\scalebox{0.73}{%
\begin{tikzpicture}[/pgfplots/width=1.5\linewidth, /pgfplots/height=1.2\linewidth]
    \begin{axis}[ymin=5,ymax=70,xmin=1,xmax=100,enlargelimits=false,
        xlabel=Confidence percentile (\%),
        ylabel=Classification accuracy (\%),
        font=\scriptsize, grid=both,
        grid style=dotted,
        axis equal image=false,
		legend cell align=left,
        legend style={at={(0.25,0.16)},anchor=south west},       
        ytick={0,10,...,100},
        xtick={0,10,...,100},
        minor ytick={5,10,...,100},
        minor xtick={5,10,...,100},
		major grid style={white!20!black},
        minor grid style={white!70!black},
        xlabel shift={-3pt},
        ylabel shift={-4pt},
		]
	    
        \addplot+[black,solid,mark=none, ultra thick]                          table[x=Percentile,y expr=100-\thisrow{Trained}] {data/orient/per_class_err.txt};
        \addlegendentry{COB (Ours)}

        \addplot+[red,solid,mark=none, ultra thick]                          table[x=Percentile,y expr=100-\thisrow{Local}] {data/orient/per_class_err.txt};
        \addlegendentry{Local gradients~\cite{Arb+11,DoZi13,Pont-Tuset2016}}
        
        \addplot+[blue,dashed,mark=none, ultra thick]                          table[x=Percentile,y expr=100-\thisrow{Random}] {data/orient/per_class_err.txt};
        \addlegendentry{Random}
	\end{axis}
   \end{tikzpicture}}
   \vspace{-7mm}
\caption{\textbf{Contour orientation}: Classification accuracy into orientations quantized in 8 bins.}
\label{fig:orient_eval}
\end{minipage}
\end{figure}

\subsection{Contour Orientation}
\label{sec:exp:orient}
We evaluate contour orientation results by the classification accuracy into 8 different orientations, to isolate their performance from the global system.
We compute the ground-truth orientations as depicted in Figure~\ref{fig:pol_simpl} by means of the sparse boundaries representation.
We then sweep all ground-truth boundary pixels and compare the estimated 
orientation with the ground-truth one.
Since the orientations are not well-balanced classes (much more horizontal
and vertical contours), we compute the classification accuracy per each of the 8 classes and then compute the mean.

Figure~\ref{fig:orient_eval} shows the classification accuracy with respect to
the confidence of the estimation.
We compare our proposed technique against the local gradient estimation used in previous literature~\cite{Arb+11,DoZi13,Pont-Tuset2016}. As a baseline, we plot the result a random guess of the orientations would get.
We observe that our estimation is significantly better than the previous approach.
As a summary measure, we compute the area under the curve of the accuracy (ours 58.6\%, local gradients 41.2\%, random 12.5\%), which corroborates the superior results from our technique.

\subsection{Generic Image Segmentation}
\label{sec:exp:generic}
We present our results for contour detection and generic image segmentation on PASCAL Context~\cite{Mot+14} as well as on the BSDS500~\cite{Martin2001}, which is the most established benchmark for perceptual grouping.

\paragraph*{\textbf{PASCAL Context:}}
We train COB in the \emph{VOC train}, and perform hyper-parameter selection on \emph{VOC val}. We report the final results on the unseen \emph{VOC test} when trained on \emph{VOC trainval}, using the previously tuned hyper-parameters. We compare our approach to several methods trained on the BSDS~\cite{DoZi13,Pont-Tuset2016,Zhao2015,XiTu15} and we also retrain the current state-of-the-art contour detection methods HED~\cite{XiTu15} and the recent CEDN~\cite{Yan+16} on \emph{VOC trainval} using the code provided by the respective authors.

Figure~\ref{fig:pr_pascal_test} presents the evaluation results of our method compared to the state-of-the-art, which show that COB outperforms all others by a considerable margin both in terms of boundaries and in terms of regions. The lower performance of the methods trained on the BSDS quantifies the difficulty of the task when moving to a larger and more challenging dataset.


\newcommand{\relplotsize}{0.8} 
\newcommand{\deflinewidth}{1.3pt} 

\begin{figure}[h!]
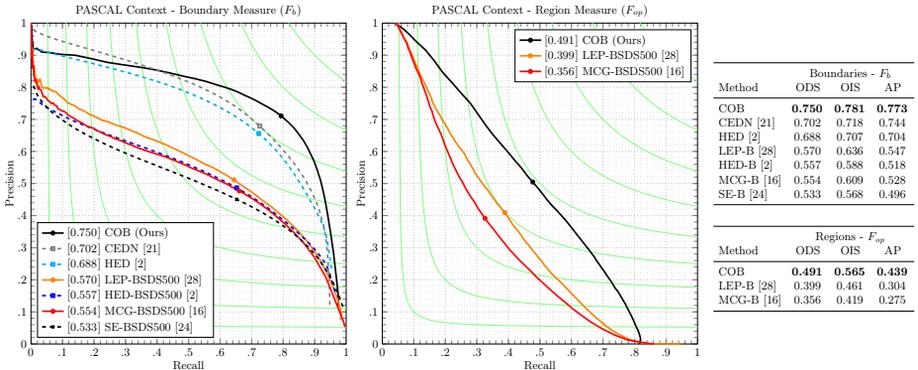

\begin{minipage}{0.77\textwidth}
\resizebox{0.5\linewidth}{!}{%
\begin{tikzpicture}[/pgfplots/width=1.1\linewidth, /pgfplots/height=1.1\linewidth]
    \begin{axis}[
                 ymin=0,ymax=1,xmin=0,xmax=1,
        		 xlabel=Recall,
        		 ylabel=Precision,
         		 xlabel shift={-2pt},
        		 ylabel shift={-3pt},
		         font=\small,
		         axis equal image=true,
		         enlargelimits=false,
		         clip=true,
        	     grid style=dotted, grid=both,
                 major grid style={white!65!black},
        		 minor grid style={white!85!black},
		 		 xtick={0,0.1,...,1.1},
        		 ytick={0,0.1,...,1.1},
         		 minor xtick={0,0.02,...,1},
		         minor ytick={0,0.02,...,1},
		         xticklabels={0,.1,.2,.3,.4,.5,.6,.7,.8,.9,1},
		         yticklabels={0,.1,.2,.3,.4,.5,.6,.7,.8,.9,1},
				 legend cell align=left,
        		 legend style={at={(0.02,0.02)},anchor=south west},
		         title style={yshift=-1ex,},
		         title={PASCAL Context - Boundary Measure ($F_b$)}]
        
    \foreach \f in {0.1,0.2,...,0.9}{%
       \addplot[white!50!green,line width=0.2pt,domain=(\f/(2-\f)):1,samples=200,forget plot]{(\f*x)/(2*x-\f)};
    }
    
    \addplot+[black,solid,mark=none, line width=\deflinewidth,forget plot] table[x=Recall,y=Precision] {data/pr/PASCALContext_test_new_fb_COB_trainval.txt};
    \addplot+[black,solid,mark=o, mark size=1.3, mark options={solid},line width=\deflinewidth] table[x=Recall,y=Precision] {data/pr/PASCALContext_test_new_fb_COB_trainval_ods.txt};
    \addlegendentry{[\showodsf{PASCALContext_test_new}{fb}{COB_trainval}] COB (Ours)}
       
    \addplot+[gray,dashed,mark=none, line width=\deflinewidth,forget plot] table[x=Recall,y=Precision] {data/pr/PASCALContext_test_new_fb_CEDN.txt};
    \addplot+[gray,dashed,mark=square,mark options={solid}, mark size=1.25, line width=\deflinewidth] table[x=Recall,y=Precision] {data/pr/PASCALContext_test_new_fb_CEDN_ods.txt};
    \addlegendentry{[\showodsf{PASCALContext_test_new}{fb}{CEDN}] CEDN~\cite{Yan+16}}
     
    \addplot+[cyan,dashed,mark=none, line width=\deflinewidth,forget plot] table[x=Recall,y=Precision] {data/pr/PASCALContext_test_new_fb_HED_trainval.txt};
    \addplot+[cyan,dashed,mark=square,mark options={solid}, mark size=1.25, line width=\deflinewidth] table[x=Recall,y=Precision] {data/pr/PASCALContext_test_new_fb_HED_trainval_ods.txt};
    \addlegendentry{[\showodsf{PASCALContext_test_new}{fb}{HED_trainval}] HED~\cite{XiTu15}}
    
    \addplot+[orange,solid,mark=none, line width=\deflinewidth,forget plot] table[x=Recall,y=Precision] {data/pr/PASCALContext_test_new_fb_LEP-BSDS500.txt};
    \addplot+[orange,solid,mark=o, mark size=1.3, mark options={solid},line width=\deflinewidth] table[x=Recall,y=Precision] {data/pr/PASCALContext_test_new_fb_LEP-BSDS500_ods.txt};
    \addlegendentry{[\showodsf{PASCALContext_test_new}{fb}{LEP-BSDS500}] LEP-BSDS500~\cite{Zhao2015}}
     
    \addplot+[blue,dashed,mark=none, line width=\deflinewidth,forget plot] table[x=Recall,y=Precision] {data/pr/PASCALContext_test_new_fb_HED-BSDS500.txt};
    \addplot+[blue,dashed,mark=square, mark size=1.25,mark options={solid},line width=\deflinewidth] table[x=Recall,y=Precision] {data/pr/PASCALContext_test_new_fb_HED-BSDS500_ods.txt};
    \addlegendentry{[\showodsf{PASCALContext_test_new}{fb}{HED-BSDS500}] HED-BSDS500~\cite{XiTu15}}
    
    \addplot+[red,solid,mark=none, line width=\deflinewidth,forget plot] table[x=Recall,y=Precision] {data/pr/PASCALContext_test_new_fb_MCG-BSDS500.txt};
    \addplot+[red,solid,mark=o, mark size=1.3, mark options={solid},line width=\deflinewidth] table[x=Recall,y=Precision] {data/pr/PASCALContext_test_new_fb_MCG-BSDS500_ods.txt};
    \addlegendentry{[\showodsf{PASCALContext_test_new}{fb}{MCG-BSDS500}] MCG-BSDS500~\cite{Pont-Tuset2016}}
    
    \addplot+[black,dashed,mark=none,line width=\deflinewidth,forget plot] table[x=Recall,y=Precision] {data/pr/PASCALContext_test_new_fb_SE-BSDS500.txt};
    \addplot+[black,dashed,mark=x, mark size=1.6, mark options={solid},line width=\deflinewidth] table[x=Recall,y=Precision] {data/pr/PASCALContext_test_new_fb_SE-BSDS500_ods.txt};
    \addlegendentry{[\showodsf{PASCALContext_test_new}{fb}{SE-BSDS500}] SE-BSDS500~\cite{DoZi13}}

    \end{axis}
\end{tikzpicture}}
\hspace{-1.5mm}
\resizebox{0.5\linewidth}{!}{%
\begin{tikzpicture}[/pgfplots/width=1.1\linewidth, /pgfplots/height=1.1\linewidth]
    \begin{axis}[
                 ymin=0,ymax=1,xmin=0,xmax=1,
        		 xlabel=Recall,
        		 ylabel=Precision,
         		 xlabel shift={-2pt},
        		 ylabel shift={-3pt},
		         font=\small,
		         axis equal image=true,
		         enlargelimits=false,
		         clip=true,
        	     grid style=dotted, grid=both,
                 major grid style={white!65!black},
        		 minor grid style={white!85!black},
		 		 xtick={0,0.1,...,1.1},
        		 ytick={0,0.1,...,1.1},
         		 minor xtick={0,0.02,...,1},
		         minor ytick={0,0.02,...,1},
		         xticklabels={0,.1,.2,.3,.4,.5,.6,.7,.8,.9,1},
		         yticklabels={0,.1,.2,.3,.4,.5,.6,.7,.8,.9,1},
				 legend cell align=left,
        		 legend style={at={(0.98,0.98)}, anchor=north east},
		         title style={yshift=-1ex,},
		         title={PASCAL Context - Region Measure ($F_{op}$)}]
        
    \foreach \f in {0.1,0.2,...,0.9}{%
       \addplot[white!50!green,line width=0.2pt,domain=(\f/(2-\f)):1,samples=200,forget plot]{(\f*x)/(2*x-\f)};
    }
	
    \addplot+[black,solid,mark=none, line width=\deflinewidth,forget plot] table[x=Recall,y=Precision] {data/pr/PASCALContext_test_new_fop_COB_trainval.txt};
    \addplot+[black,solid,mark=o, mark size=1.3, mark options={solid},line width=\deflinewidth] table[x=Recall,y=Precision] {data/pr/PASCALContext_test_new_fop_COB_trainval_ods.txt};
    \addlegendentry{[\showodsf{PASCALContext_test_new}{fop}{COB_trainval}] COB (Ours)}
    
    \addplot+[orange,solid,mark=none, line width=\deflinewidth,forget plot] table[x=Recall,y=Precision] {data/pr/PASCALContext_test_new_fop_LEP-BSDS500.txt};
    \addplot+[orange,solid,mark=o, mark size=1.3, mark options={solid},line width=\deflinewidth] table[x=Recall,y=Precision] {data/pr/PASCALContext_test_new_fop_LEP-BSDS500_ods.txt};
    \addlegendentry{[\showodsf{PASCALContext_test_new}{fop}{LEP-BSDS500}] LEP-BSDS500~\cite{Zhao2015}}
	
    \addplot+[red,solid,mark=none, line width=\deflinewidth,forget plot] table[x=Recall,y=Precision] {data/pr/PASCALContext_test_new_fop_MCG-BSDS500.txt};
    \addplot+[red,solid,mark=o, mark size=1.3, mark options={solid},line width=\deflinewidth] table[x=Recall,y=Precision] {data/pr/PASCALContext_test_new_fop_MCG-BSDS500_ods.txt};
    \addlegendentry{[\showodsf{PASCALContext_test_new}{fop}{MCG-BSDS500}] MCG-BSDS500~\cite{Pont-Tuset2016}}

    \end{axis}
\end{tikzpicture}}
\end{minipage}
\begin{minipage}{0.22\linewidth}
\setlength{\tabcolsep}{4pt} 
\center
\footnotesize
\resizebox{\textwidth}{!}{%
\begin{tabular}{lccc}
\toprule
       & \multicolumn{3}{c}{Boundaries - $F_b$} \\
Method &  ODS & OIS & AP\\
\midrule
COB   & \bf%
\mbox{\input{data/pr/PASCALContext_test_new_fb_COB_trainval_ods_f.txt}\hspace{-2.5pt}}%
 & \bf%
\mbox{\input{data/pr/PASCALContext_test_new_fb_COB_trainval_ois_f.txt}\hspace{-2.5pt}}%
 & \bf%
\mbox{\input{data/pr/PASCALContext_test_new_fb_COB_trainval_ap.txt}\hspace{-2.5pt}}%
 \\
CEDN~\cite{Yan+16}   & %
\mbox{\input{data/pr/PASCALContext_test_new_fb_CEDN_ods_f.txt}\hspace{-2.5pt}}%
 & %
\mbox{\input{data/pr/PASCALContext_test_new_fb_CEDN_ois_f.txt}\hspace{-2.5pt}}%
 & %
\mbox{\input{data/pr/PASCALContext_test_new_fb_CEDN_ap.txt}\hspace{-2.5pt}}%
 \\
HED~\cite{XiTu15}   & %
\mbox{\input{data/pr/PASCALContext_test_new_fb_HED_trainval_ods_f.txt}\hspace{-2.5pt}}%
 & %
\mbox{\input{data/pr/PASCALContext_test_new_fb_HED_trainval_ois_f.txt}\hspace{-2.5pt}}%
 & %
\mbox{\input{data/pr/PASCALContext_test_new_fb_HED_trainval_ap.txt}\hspace{-2.5pt}}%
 \\
LEP-B~\cite{Zhao2015}   & %
\mbox{\input{data/pr/PASCALContext_test_new_fb_LEP-BSDS500_ods_f.txt}\hspace{-2.5pt}}%
 & %
\mbox{\input{data/pr/PASCALContext_test_new_fb_LEP-BSDS500_ois_f.txt}\hspace{-2.5pt}}%
 & %
\mbox{\input{data/pr/PASCALContext_test_new_fb_LEP-BSDS500_ap.txt}\hspace{-2.5pt}}%
 \\
HED-B~\cite{XiTu15}   & %
\mbox{\input{data/pr/PASCALContext_test_new_fb_HED-BSDS500_ods_f.txt}\hspace{-2.5pt}}%
 & %
\mbox{\input{data/pr/PASCALContext_test_new_fb_HED-BSDS500_ois_f.txt}\hspace{-2.5pt}}%
 & %
\mbox{\input{data/pr/PASCALContext_test_new_fb_HED-BSDS500_ap.txt}\hspace{-2.5pt}}%
 \\
MCG-B~\cite{Pont-Tuset2016}   & %
\mbox{\input{data/pr/PASCALContext_test_new_fb_MCG-BSDS500_ods_f.txt}\hspace{-2.5pt}}%
 & %
\mbox{\input{data/pr/PASCALContext_test_new_fb_MCG-BSDS500_ois_f.txt}\hspace{-2.5pt}}%
 & %
\mbox{\input{data/pr/PASCALContext_test_new_fb_MCG-BSDS500_ap.txt}\hspace{-2.5pt}}%
 \\
SE-B~\cite{DoZi13}   & %
\mbox{\input{data/pr/PASCALContext_test_new_fb_SE-BSDS500_ods_f.txt}\hspace{-2.5pt}}%
 & %
\mbox{\input{data/pr/PASCALContext_test_new_fb_SE-BSDS500_ois_f.txt}\hspace{-2.5pt}}%
 & %
\mbox{\input{data/pr/PASCALContext_test_new_fb_SE-BSDS500_ap.txt}\hspace{-2.5pt}}%
 \\
\bottomrule
\end{tabular}}
\rule{0mm}{2mm}
\resizebox{\textwidth}{!}{%
\begin{tabular}{lccc}
\toprule
       & \multicolumn{3}{c}{Regions - $F_\mathit{op}$} \\
Method &  ODS & OIS & AP\\
\midrule
COB   & \bf%
\mbox{\input{data/pr/PASCALContext_test_new_fop_COB_trainval_ods_f.txt}\hspace{-2.5pt}}%
 & \bf%
\mbox{\input{data/pr/PASCALContext_test_new_fop_COB_trainval_ois_f.txt}\hspace{-2.5pt}}%
 & \bf%
\mbox{\input{data/pr/PASCALContext_test_new_fop_COB_trainval_ap.txt}\hspace{-2.5pt}}%
 \\
LEP-B~\cite{Zhao2015}   & %
\mbox{\input{data/pr/PASCALContext_test_new_fop_LEP-BSDS500_ods_f.txt}\hspace{-2.5pt}}%
 & %
\mbox{\input{data/pr/PASCALContext_test_new_fop_LEP-BSDS500_ois_f.txt}\hspace{-2.5pt}}%
 & %
\mbox{\input{data/pr/PASCALContext_test_new_fop_LEP-BSDS500_ap.txt}\hspace{-2.5pt}}%
 \\
MCG-B~\cite{Pont-Tuset2016}   & %
\mbox{\input{data/pr/PASCALContext_test_new_fop_MCG-BSDS500_ods_f.txt}\hspace{-2.5pt}}%
 & %
\mbox{\input{data/pr/PASCALContext_test_new_fop_MCG-BSDS500_ois_f.txt}\hspace{-2.5pt}}%
 & %
\mbox{\input{data/pr/PASCALContext_test_new_fop_MCG-BSDS500_ap.txt}\hspace{-2.5pt}}%
 \\
\bottomrule
\end{tabular}}
\end{minipage}
\vspace{-3mm}
\caption{\textbf{PASCAL Context \textit{VOC test} Evaluation}: Precision-recall curves for evaluation of boundaries ($F_b$~\cite{Martin2004}), and regions ($F_{op}$~\cite{Pont-Tuset2016a}). Contours in dashed lines and boundaries (from segmentation) in solid lines. ODS, OIS, and AP summary measures.}
\label{fig:pr_pascal_test}
\end{figure}

\paragraph*{\textbf{BSDS500:}}
We retrain COB using only the 300 images of the \textit{trainval} set of the BSDS, after data augmentation as suggested in~\cite{XiTu15}, keeping the architecture decided in Section~\ref{sec:exp:ablat}. For comparison to HED~\cite{XiTu15}, we used the model that the authors provide online. We also compare with CEDN~\cite{Yan+16}, by evaluating the results provided by the authors. 

\begin{figure}[h!]
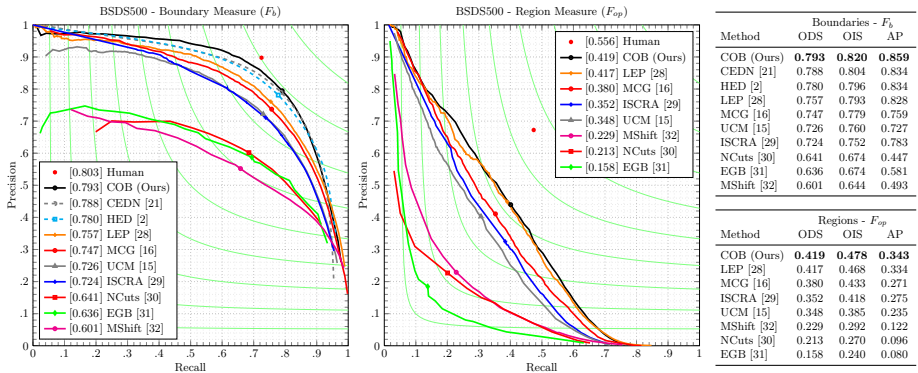

\begin{minipage}{0.77\textwidth}
\resizebox{0.5\linewidth}{!}{%
\begin{tikzpicture}[/pgfplots/width=1.1\linewidth, /pgfplots/height=1.1\linewidth]
    \begin{axis}[
                 ymin=0,ymax=1,xmin=0,xmax=1,
        		 xlabel=Recall,
        		 ylabel=Precision,
         		 xlabel shift={-2pt},
        		 ylabel shift={-3pt},
		         font=\small,
		         axis equal image=true,
		         enlargelimits=false,
		         clip=true,
        	     grid style=dotted, grid=both,
                 major grid style={white!65!black},
        		 minor grid style={white!85!black},
		 		 xtick={0,0.1,...,1.1},
        		 ytick={0,0.1,...,1.1},
         		 minor xtick={0,0.02,...,1},
		         minor ytick={0,0.02,...,1},
		         xticklabels={0,.1,.2,.3,.4,.5,.6,.7,.8,.9,1},
		         yticklabels={0,.1,.2,.3,.4,.5,.6,.7,.8,.9,1},
				 legend cell align=left,
        		 legend style={at={(0.02,0.02)},anchor=south west},
		         title style={yshift=-1ex,},
		         title={BSDS500 - Boundary Measure ($F_b$)}]
        
    \foreach \f in {0.1,0.2,...,0.9}{%
       \addplot[white!50!green,line width=0.2pt,domain=(\f/(2-\f)):1,samples=200,forget plot]{(\f*x)/(2*x-\f)};
    }

    \addplot+[only marks,red,mark=asterisk,mark size=1.7,line width=\deflinewidth] table[x=Recall,y=Precision] {data/pr/BSDS500_test_fb_human.txt};
    \addlegendentry{[\showodsf{BSDS500_test}{fb}{human}] Human}
	
    \addplot+[black,solid,mark=none, line width=\deflinewidth,forget plot] table[x=Recall,y=Precision] {data/pr/BSDS500_test_fb_COB.txt};
    \addplot+[black,solid,mark=o, mark size=1.3, line width=\deflinewidth] table[x=Recall,y=Precision] {data/pr/BSDS500_test_fb_COB_ods.txt};
    \addlegendentry{[\showodsf{BSDS500_test}{fb}{COB}] COB (Ours)}
    
    \addplot+[gray,dashed,mark=none, line width=\deflinewidth,forget plot] table[x=Recall,y=Precision] {data/pr/BSDS500_test_fb_CEDN.txt};
    \addplot+[gray,dashed,mark=o, mark size=1.3, line width=\deflinewidth] table[x=Recall,y=Precision] {data/pr/BSDS500_test_fb_CEDN_ods.txt};
    \addlegendentry{[\showodsf{BSDS500_test}{fb}{CEDN}] CEDN~\cite{Yan+16}}

    \addplot+[cyan,dashed,mark=none, line width=\deflinewidth,forget plot] table[x=Recall,y=Precision] {data/pr/BSDS500_test_fb_HED.txt};
    \addplot+[cyan,dashed,mark=square,mark options={solid}, mark size=1.25, line width=\deflinewidth] table[x=Recall,y=Precision] {data/pr/BSDS500_test_fb_HED_ods.txt};
    \addlegendentry{[\showodsf{BSDS500_test}{fb}{HED}] HED~\cite{XiTu15}}
    
    \addplot+[orange,solid,mark=none, line width=\deflinewidth,forget plot] table[x=Recall,y=Precision] {data/pr/BSDS500_test_fb_LEP.txt};
    \addplot+[orange,solid,mark=+, mark size=1.6, line width=\deflinewidth] table[x=Recall,y=Precision] {data/pr/BSDS500_test_fb_LEP_ods.txt};
    \addlegendentry{[\showodsf{BSDS500_test}{fb}{LEP}] LEP~\cite{Zhao2015}}
    
    \addplot+[red,solid,mark=none, line width=\deflinewidth,forget plot] table[x=Recall,y=Precision] {data/pr/BSDS500_test_fb_MCG.txt};
    \addplot+[red,solid,mark=o, mark size=1.3, line width=\deflinewidth] table[x=Recall,y=Precision] {data/pr/BSDS500_test_fb_MCG_ods.txt};
    \addlegendentry{[\showodsf{BSDS500_test}{fb}{MCG}] MCG~\cite{Pont-Tuset2016}}

    \addplot+[gray,solid,mark=none, line width=\deflinewidth,forget plot] table[x=Recall,y=Precision] {data/pr/BSDS500_test_fb_gPb-UCM.txt};
    \addplot+[gray,solid,mark=triangle, mark size=1.6, line width=\deflinewidth] table[x=Recall,y=Precision] {data/pr/BSDS500_test_fb_gPb-UCM_ods.txt};
    \addlegendentry{[\showodsf{BSDS500_test}{fb}{gPb-UCM}] UCM~\cite{Arb+11}}
    
    \addplot+[blue,solid,mark=none,  line width=\deflinewidth,forget plot] table[x=Recall,y=Precision] {data/pr/BSDS500_test_fb_ISCRA.txt};
    \addplot+[blue,solid,mark=+, mark size=1.6,  line width=\deflinewidth] table[x=Recall,y=Precision] {data/pr/BSDS500_test_fb_ISCRA_ods.txt};
    \addlegendentry{[\showodsf{BSDS500_test}{fb}{ISCRA}] ISCRA~\cite{Ren2013}}
	
    \addplot+[red,solid,mark=none,  line width=\deflinewidth,forget plot] table[x=Recall,y=Precision] {data/pr/BSDS500_test_fb_NCut.txt};
    \addplot+[red,solid,mark=square, mark size=1.25,  line width=\deflinewidth] table[x=Recall,y=Precision] {data/pr/BSDS500_test_fb_NCut_ods.txt};
    \addlegendentry{[\showodsf{BSDS500_test}{fb}{NCut}] NCuts~\cite{Shi2000}}
	
    \addplot+[green,solid,mark=none,  line width=\deflinewidth,forget plot] table[x=Recall,y=Precision] {data/pr/BSDS500_test_fb_EGB.txt};
    \addplot+[green,solid,mark=diamond, mark size=1.5,  line width=\deflinewidth] table[x=Recall,y=Precision] {data/pr/BSDS500_test_fb_EGB_ods.txt};
    \addlegendentry{[\showodsf{BSDS500_test}{fb}{EGB}] EGB~\cite{Felzenszwalb2004}}
	
    \addplot+[magenta,solid,mark=none,line width=\deflinewidth,forget plot] table[x=Recall,y=Precision] {data/pr/BSDS500_test_fb_MShift.txt};
    \addplot+[magenta,solid,mark=o, mark size=1.3,  line width=\deflinewidth] table[x=Recall,y=Precision] {data/pr/BSDS500_test_fb_MShift_ods.txt};
    \addlegendentry{[\showodsf{BSDS500_test}{fb}{MShift}] MShift~\cite{Comaniciu2002}}

    \end{axis}
\end{tikzpicture}}
\hspace{-1.5mm}
\resizebox{0.5\linewidth}{!}{%
\begin{tikzpicture}[/pgfplots/width=1.1\linewidth, /pgfplots/height=1.1\linewidth]
    \begin{axis}[
                 ymin=0,ymax=1,xmin=0,xmax=1,
        		 xlabel=Recall,
        		 ylabel=Precision,
         		 xlabel shift={-2pt},
        		 ylabel shift={-3pt},
		         font=\small,
		         axis equal image=true,
		         enlargelimits=false,
		         clip=true,
        	     grid style=dotted, grid=both,
                 major grid style={white!65!black},
        		 minor grid style={white!85!black},
		 		 xtick={0,0.1,...,1.1},
        		 ytick={0,0.1,...,1.1},
         		 minor xtick={0,0.02,...,1},
		         minor ytick={0,0.02,...,1},
		         xticklabels={0,.1,.2,.3,.4,.5,.6,.7,.8,.9,1},
		         yticklabels={0,.1,.2,.3,.4,.5,.6,.7,.8,.9,1},
				 legend cell align=left,
        		 legend style={at={(0.98,0.98)},anchor=north east},
		         title style={yshift=-1ex,},
		         title={BSDS500 - Region Measure ($F_{op}$)}]
        
    \foreach \f in {0.1,0.2,...,0.9}{%
       \addplot[white!50!green,line width=0.2pt,domain=(\f/(2-\f)):1,samples=200,forget plot]{(\f*x)/(2*x-\f)};
    }
	
    \addplot+[only marks,red,mark=asterisk,mark size=1.7,line width=\deflinewidth] table[x=Recall,y=Precision] {data/pr/BSDS500_test_fop_human.txt};
    \addlegendentry{[\showodsf{BSDS500_test}{fop}{human}] Human}
	
    \addplot+[black,solid,mark=none, line width=\deflinewidth,forget plot] table[x=Recall,y=Precision] {data/pr/BSDS500_test_fop_COB.txt};
    \addplot+[black,solid,mark=o, mark size=1.3, line width=\deflinewidth] table[x=Recall,y=Precision] {data/pr/BSDS500_test_fop_COB_ods.txt};
    \addlegendentry{[\showodsf{BSDS500_test}{fop}{COB}] COB (Ours)}
    
    \addplot+[orange,solid,mark=none, line width=\deflinewidth,forget plot] table[x=Recall,y=Precision] {data/pr/BSDS500_test_fop_LEP.txt};
    \addplot+[orange,solid,mark=+, mark size=1.6, line width=\deflinewidth] table[x=Recall,y=Precision] {data/pr/BSDS500_test_fop_LEP_ods.txt};
    \addlegendentry{[\showodsf{BSDS500_test}{fop}{LEP}] LEP~\cite{Zhao2015}}

    \addplot+[red,solid,mark=none, line width=\deflinewidth,forget plot] table[x=Recall,y=Precision] {data/pr/BSDS500_test_fop_MCG.txt};
    \addplot+[red,solid,mark=o, mark size=1.3, line width=\deflinewidth] table[x=Recall,y=Precision] {data/pr/BSDS500_test_fop_MCG_ods.txt};
    \addlegendentry{[\showodsf{BSDS500_test}{fop}{MCG}] MCG~\cite{Pont-Tuset2016}}
    
    \addplot+[blue,solid,mark=none,  line width=\deflinewidth,forget plot] table[x=Recall,y=Precision] {data/pr/BSDS500_test_fop_ISCRA.txt};
    \addplot+[blue,solid,mark=+, mark size=1.6,  line width=\deflinewidth] table[x=Recall,y=Precision] {data/pr/BSDS500_test_fop_ISCRA_ods.txt};
    \addlegendentry{[\showodsf{BSDS500_test}{fop}{ISCRA}] ISCRA~\cite{Ren2013}}
    
    \addplot+[gray,solid,mark=none, line width=\deflinewidth,forget plot] table[x=Recall,y=Precision] {data/pr/BSDS500_test_fop_gPb-UCM.txt};
    \addplot+[gray,solid,mark=triangle, mark size=1.6, line width=\deflinewidth] table[x=Recall,y=Precision] {data/pr/BSDS500_test_fop_gPb-UCM_ods.txt};
    \addlegendentry{[\showodsf{BSDS500_test}{fop}{gPb-UCM}] UCM~\cite{Arb+11}}
   
    \addplot+[magenta,solid,mark=none,line width=\deflinewidth,forget plot] table[x=Recall,y=Precision] {data/pr/BSDS500_test_fop_MShift.txt};
    \addplot+[magenta,solid,mark=o, mark size=1.3,  line width=\deflinewidth] table[x=Recall,y=Precision] {data/pr/BSDS500_test_fop_MShift_ods.txt};
    \addlegendentry{[\showodsf{BSDS500_test}{fop}{MShift}] MShift~\cite{Comaniciu2002}}
	
    \addplot+[red,solid,mark=none,  line width=\deflinewidth,forget plot] table[x=Recall,y=Precision] {data/pr/BSDS500_test_fop_NCut.txt};
    \addplot+[red,solid,mark=square, mark size=1.25,  line width=\deflinewidth] table[x=Recall,y=Precision] {data/pr/BSDS500_test_fop_NCut_ods.txt};
    \addlegendentry{[\showodsf{BSDS500_test}{fop}{NCut}] NCuts~\cite{Shi2000}}
	
    \addplot+[green,solid,mark=none,  line width=\deflinewidth,forget plot] table[x=Recall,y=Precision] {data/pr/BSDS500_test_fop_EGB.txt};
    \addplot+[green,solid,mark=diamond, mark size=1.5,  line width=\deflinewidth] table[x=Recall,y=Precision] {data/pr/BSDS500_test_fop_EGB_ods.txt};
    \addlegendentry{[\showodsf{BSDS500_test}{fop}{EGB}] EGB~\cite{Felzenszwalb2004}}

    \end{axis}
\end{tikzpicture}}
\end{minipage}
\begin{minipage}{0.22\linewidth}
\setlength{\tabcolsep}{4pt} 
\center
\footnotesize
\resizebox{\textwidth}{!}{%
\begin{tabular}{lccc}
\toprule
       & \multicolumn{3}{c}{Boundaries - $F_b$} \\
Method &  ODS & OIS & AP\\
\midrule
COB (Ours)  & \bf%
\mbox{\input{data/pr/BSDS500_test_fb_COB_ods_f.txt}\hspace{-2.5pt}}%
 & \bf%
\mbox{\input{data/pr/BSDS500_test_fb_COB_ois_f.txt}\hspace{-2.5pt}}%
 & \bf%
\mbox{\input{data/pr/BSDS500_test_fb_COB_ap.txt}\hspace{-2.5pt}}%
 \\
CEDN~\cite{Yan+16}  & %
\mbox{\input{data/pr/BSDS500_test_fb_CEDN_ods_f.txt}\hspace{-2.5pt}}%
 & %
\mbox{\input{data/pr/BSDS500_test_fb_CEDN_ois_f.txt}\hspace{-2.5pt}}%
 & %
\mbox{\input{data/pr/BSDS500_test_fb_CEDN_ap.txt}\hspace{-2.5pt}}%
 \\
HED~\cite{XiTu15}   & %
\mbox{\input{data/pr/BSDS500_test_fb_HED_ods_f.txt}\hspace{-2.5pt}}%
 & %
\mbox{\input{data/pr/BSDS500_test_fb_HED_ois_f.txt}\hspace{-2.5pt}}%
 & %
\mbox{\input{data/pr/BSDS500_test_fb_HED_ap.txt}\hspace{-2.5pt}}%
 \\
LEP~\cite{Zhao2015}   & %
\mbox{\input{data/pr/BSDS500_test_fb_LEP_ods_f.txt}\hspace{-2.5pt}}%
 & %
\mbox{\input{data/pr/BSDS500_test_fb_LEP_ois_f.txt}\hspace{-2.5pt}}%
 & %
\mbox{\input{data/pr/BSDS500_test_fb_LEP_ap.txt}\hspace{-2.5pt}}%
 \\
MCG~\cite{Pont-Tuset2016}   & %
\mbox{\input{data/pr/BSDS500_test_fb_MCG_ods_f.txt}\hspace{-2.5pt}}%
 &  %
\mbox{\input{data/pr/BSDS500_test_fb_MCG_ois_f.txt}\hspace{-2.5pt}}%
 & %
\mbox{\input{data/pr/BSDS500_test_fb_MCG_ap.txt}\hspace{-2.5pt}}%
 \\
UCM~\cite{Arb+11}   & %
\mbox{\input{data/pr/BSDS500_test_fb_gPb-UCM_ods_f.txt}\hspace{-2.5pt}}%
 &  %
\mbox{\input{data/pr/BSDS500_test_fb_gPb-UCM_ois_f.txt}\hspace{-2.5pt}}%
 & %
\mbox{\input{data/pr/BSDS500_test_fb_gPb-UCM_ap.txt}\hspace{-2.5pt}}%
 \\
ISCRA~\cite{Ren2013}   & %
\mbox{\input{data/pr/BSDS500_test_fb_ISCRA_ods_f.txt}\hspace{-2.5pt}}%
 & %
\mbox{\input{data/pr/BSDS500_test_fb_ISCRA_ois_f.txt}\hspace{-2.5pt}}%
 & %
\mbox{\input{data/pr/BSDS500_test_fb_ISCRA_ap.txt}\hspace{-2.5pt}}%
 \\
NCuts~\cite{Shi2000}   & %
\mbox{\input{data/pr/BSDS500_test_fb_NCut_ods_f.txt}\hspace{-2.5pt}}%
 &  %
\mbox{\input{data/pr/BSDS500_test_fb_NCut_ois_f.txt}\hspace{-2.5pt}}%
 & %
\mbox{\input{data/pr/BSDS500_test_fb_NCut_ap.txt}\hspace{-2.5pt}}%
 \\
EGB~\cite{Felzenszwalb2004}   & %
\mbox{\input{data/pr/BSDS500_test_fb_EGB_ods_f.txt}\hspace{-2.5pt}}%
 &  %
\mbox{\input{data/pr/BSDS500_test_fb_EGB_ois_f.txt}\hspace{-2.5pt}}%
 & %
\mbox{\input{data/pr/BSDS500_test_fb_EGB_ap.txt}\hspace{-2.5pt}}%
 \\
MShift~\cite{Comaniciu2002}   & %
\mbox{\input{data/pr/BSDS500_test_fb_MShift_ods_f.txt}\hspace{-2.5pt}}%
 &  %
\mbox{\input{data/pr/BSDS500_test_fb_MShift_ois_f.txt}\hspace{-2.5pt}}%
 & %
\mbox{\input{data/pr/BSDS500_test_fb_MShift_ap.txt}\hspace{-2.5pt}}%
 \\
\bottomrule
\end{tabular}}
\rule{0mm}{1mm}
\resizebox{\textwidth}{!}{%
\begin{tabular}{lccc}
\toprule
       & \multicolumn{3}{c}{Regions - $F_\mathit{op}$} \\
Method &  ODS & OIS & AP\\
\midrule
COB (Ours)  & \bf%
\mbox{\input{data/pr/BSDS500_test_fop_COB_ods_f.txt}\hspace{-2.5pt}}%
 & \bf%
\mbox{\input{data/pr/BSDS500_test_fop_COB_ois_f.txt}\hspace{-2.5pt}}%
 & \bf%
\mbox{\input{data/pr/BSDS500_test_fop_COB_ap.txt}\hspace{-2.5pt}}%
 \\
LEP~\cite{Zhao2015}   & %
\mbox{\input{data/pr/BSDS500_test_fop_LEP_ods_f.txt}\hspace{-2.5pt}}%
 & %
\mbox{\input{data/pr/BSDS500_test_fop_LEP_ois_f.txt}\hspace{-2.5pt}}%
 & %
\mbox{\input{data/pr/BSDS500_test_fop_LEP_ap.txt}\hspace{-2.5pt}}%
 \\
MCG~\cite{Pont-Tuset2016}   & %
\mbox{\input{data/pr/BSDS500_test_fop_MCG_ods_f.txt}\hspace{-2.5pt}}%
 & %
\mbox{\input{data/pr/BSDS500_test_fop_MCG_ois_f.txt}\hspace{-2.5pt}}%
 & %
\mbox{\input{data/pr/BSDS500_test_fop_MCG_ap.txt}\hspace{-2.5pt}}%
 \\
ISCRA~\cite{Ren2013}   & %
\mbox{\input{data/pr/BSDS500_test_fop_ISCRA_ods_f.txt}\hspace{-2.5pt}}%
 &  %
\mbox{\input{data/pr/BSDS500_test_fop_ISCRA_ois_f.txt}\hspace{-2.5pt}}%
 & %
\mbox{\input{data/pr/BSDS500_test_fop_ISCRA_ap.txt}\hspace{-2.5pt}}%
 \\
UCM~\cite{Arb+11}   & %
\mbox{\input{data/pr/BSDS500_test_fop_gPb-UCM_ods_f.txt}\hspace{-2.5pt}}%
 &  %
\mbox{\input{data/pr/BSDS500_test_fop_gPb-UCM_ois_f.txt}\hspace{-2.5pt}}%
 & %
\mbox{\input{data/pr/BSDS500_test_fop_gPb-UCM_ap.txt}\hspace{-2.5pt}}%
 \\
MShift~\cite{Comaniciu2002}   & %
\mbox{\input{data/pr/BSDS500_test_fop_MShift_ods_f.txt}\hspace{-2.5pt}}%
 &  %
\mbox{\input{data/pr/BSDS500_test_fop_MShift_ois_f.txt}\hspace{-2.5pt}}%
 & %
\mbox{\input{data/pr/BSDS500_test_fop_MShift_ap.txt}\hspace{-2.5pt}}%
 \\
NCuts~\cite{Shi2000}   & %
\mbox{\input{data/pr/BSDS500_test_fop_NCut_ods_f.txt}\hspace{-2.5pt}}%
 &  %
\mbox{\input{data/pr/BSDS500_test_fop_NCut_ois_f.txt}\hspace{-2.5pt}}%
 & %
\mbox{\input{data/pr/BSDS500_test_fop_NCut_ap.txt}\hspace{-2.5pt}}%
 \\
EGB~\cite{Felzenszwalb2004}   & %
\mbox{\input{data/pr/BSDS500_test_fop_EGB_ods_f.txt}\hspace{-2.5pt}}%
 &  %
\mbox{\input{data/pr/BSDS500_test_fop_EGB_ois_f.txt}\hspace{-2.5pt}}%
 & %
\mbox{\input{data/pr/BSDS500_test_fop_EGB_ap.txt}\hspace{-2.5pt}}%
 \\
\bottomrule
\end{tabular}}
\end{minipage}
\vspace{-3mm}
\caption{\textbf{BSDS500 Test Evaluation}: Precision-recall curves for evaluation of boundaries ($F_b$~\cite{Martin2004}), and regions ($F_{op}$~\cite{Pont-Tuset2016a}). ODS, OIS, and AP summary measures.}
\label{fig:pr_bsds_test}
\end{figure}

Figure~\ref{fig:pr_bsds_test} presents the evaluation results, which show that we also obtain state-of-the-art results in this dataset. The smaller margins are in all likelihood due to the fact that we almost reach human performance for the task of contour detection on the BSDS, which motivates the shift to PASCAL Context to achieve further progress in the field.

\subsection{Object Proposals}
\label{sec:exp:prop}
Object proposals are an integral part of current object detection and semantic segmentation pipelines~\cite{Girshick2014,Girshick2015,Ren2015}, as they provide a reduced search space on locations, scales, and shapes over the image.
This section evaluates COB as a segmented proposal technique, when using our
high-quality region hierarchies in conjunction with the combinatorial grouping framework of~\cite{Pont-Tuset2016}.
We compare against the more recent techniques POISE~\cite{Humayun2015}, MCG and SCG~\cite{Pont-Tuset2016},
	    LPO~\cite{Kraehenbuehl2015}, GOP~\cite{Kraehenbuehl2014}, SeSe~\cite{Uijlings2013}, GLS~\cite{Rantalankila2014}, and RIGOR~\cite{Humayun2014}. Recent thorough comparisons of object proposal generation methods can be found in~\cite{Hosang2015,Pont-Tuset2015b}.
	     
We perform experiments on the PASCAL 2012 Segmentation dataset~\cite{Eve+12} and on the bigger and more challenging MS-COCO~\cite{Lin2014a} (val set). The hierarchies and combinatorial grouping are trained on PASCAL Context. To assess the generalization capability, we evaluate on MS-COCO, which contains a large number of previously unseen categories, without further retraining.

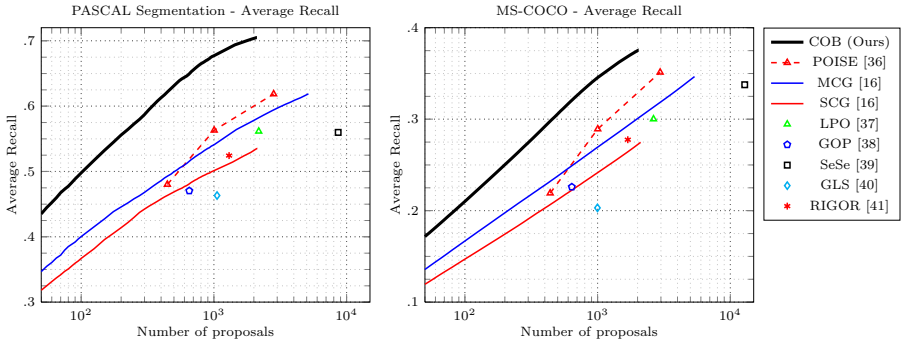
\begin{figure}[h]
\hspace{-3mm}
\begin{minipage}[b]{0.45\linewidth}
\centering
\scalebox{0.72}{%
\begin{tikzpicture}[/pgfplots/width=1.4\linewidth, /pgfplots/height=1.2\linewidth]
    \begin{axis}[ymin=0.3,ymax=0.72,xmin=50,xmax=15000,enlargelimits=false,
        xlabel=Number of proposals,
        ylabel=Average Recall,
        font=\scriptsize, grid=both,
        grid style=dotted,
        axis equal image=false,
        ytick={0,0.1,...,1},
        yticklabels={0,.1,.2,.3,.4,.5,.6,.7,.8,.9,1},
        minor ytick={0,0.025,...,1},
        major grid style={white!20!black},
        minor grid style={white!70!black},
        xlabel shift={-3pt},
        ylabel shift={-4pt},
        xmode=log,
		title style={yshift=-1ex,},
		title={PASCAL Segmentation - Average Recall}]]

		\addplot+[black,mark=none, ultra thick]                          table[x=ncands,y=average_recall] {data/obj_cands/Pascal_Segmentation_val_2012_COB.txt};
        \addplot+[red,dashed,mark=triangle, mark options={solid}, mark size=1.8, thick]                          table[x=ncands,y=average_recall] {data/obj_cands/Pascal_Segmentation_val_2012_POISE.txt};
        \addplot+[blue,solid,mark=none, thick]                          table[x=ncands,y=average_recall] {data/obj_cands/Pascal_Segmentation_val_2012_MCG.txt};
        \addplot+[red,solid,mark=none, thick]                                  table[x=ncands,y=average_recall] {data/obj_cands/Pascal_Segmentation_val_2012_SCG.txt};
        \addplot+[only marks,blue,solid,mark=pentagon,mark size=1.8, thick]    table[x=ncands,y=average_recall] {data/obj_cands/Pascal_Segmentation_val_2012_GOP.txt};
	  	\addplot+[only marks,cyan,solid,mark=diamond,mark size=1.9, thick]      table[x=ncands,y=average_recall] {data/obj_cands/Pascal_Segmentation_val_2012_GLS.txt};
	    \addplot+[only marks,red,solid,mark=asterisk, mark size=1.8, thick]   table[x=ncands,y=average_recall] {data/obj_cands/Pascal_Segmentation_val_2012_RIGOR.txt};
	    \addplot+[only marks,black,solid,mark=square, mark size=1.45, thick]    table[x=ncands,y=average_recall] {data/obj_cands/Pascal_Segmentation_val_2012_SeSe.txt};
	    \addplot+[only marks,green,solid,mark=triangle, mark size=1.8, thick]    table[x=ncands,y=average_recall] {data/obj_cands/Pascal_Segmentation_val_2012_LPO.txt};
	\end{axis}
   \end{tikzpicture}}
\end{minipage}
\hspace{-3mm}
\begin{minipage}[b]{0.45\linewidth}
\centering
\scalebox{0.72}{%
\begin{tikzpicture}[/pgfplots/width=1.4\linewidth, /pgfplots/height=1.2\linewidth]
    \begin{axis}[ymin=0.1,ymax=0.4,xmin=50,xmax=15000,enlargelimits=false,
        xlabel=Number of proposals,
        ylabel=Average Recall,
        font=\scriptsize, grid=both,
        grid style=dotted,
        axis equal image=false,
        legend pos= outer north east,
        ytick={0,0.1,...,1},
        yticklabels={0,.1,.2,.3,.4,.5,.6,.7,.8,.9,1},
        minor ytick={0,0.025,...,1},
        major grid style={white!20!black},
        minor grid style={white!70!black},
        xlabel shift={-3pt},
        ylabel shift={-4pt},
        xmode=log,
		title style={yshift=-1ex,},
		title={MS-COCO - Average Recall}]]
		
		\addplot[black,solid,mark=none, ultra thick]                          coordinates {( 1, 0.7)( 1, 0.8)};
        \label{fig:recall:ours}
        \addplot[red,dashed,mark=triangle, mark options={solid}, mark size=1.8, thick]                          coordinates {( 1, 0.7)( 1, 0.8)};
        \label{fig:recall:poise}
        \addplot[blue,solid,mark=none, thick]                          coordinates {( 1, 0.7)( 1, 0.8)};
        \label{fig:recall:mcg}
        \addplot[red,solid,mark=none, thick]                                  coordinates {( 1, 0.7)( 1, 0.8)};
        \label{fig:recall:scg}
	    \addplot[only marks,green,solid,mark=triangle, mark size=1.8, thick]                                coordinates {( 1, 0.7)( 1, 0.8)};
        \addplot[only marks,blue,solid,mark=pentagon,mark size=1.8, thick]    coordinates {( 1, 0.7)( 1, 0.8)};
        \label{fig:recall:gop}
        \addplot[only marks,black,solid,mark=square, mark size=1.45, thick]    coordinates {( 1, 0.7)( 1, 0.8)};
	    \label{fig:recall:sese}
	  	\addplot[only marks,cyan,solid,mark=diamond,mark size=1.9, thick]      coordinates {( 1, 0.7)( 1, 0.8)};
		\label{fig:recall:gls}
	    \addplot[only marks,red,solid,mark=asterisk, mark size=1.8, thick]   coordinates {( 1, 0.7)( 1, 0.8)};
	    \label{fig:recall:rigor}

        \addlegendentry{COB (Ours)}
        \addlegendentry{POISE~\cite{Humayun2015}}
		\addlegendentry{MCG~\cite{Pont-Tuset2016}}
        \addlegendentry{SCG~\cite{Pont-Tuset2016}}
	    \addlegendentry{LPO~\cite{Kraehenbuehl2015}}
        \addlegendentry{GOP~\cite{Kraehenbuehl2014}}
		\addlegendentry{SeSe~\cite{Uijlings2013}}
		\addlegendentry{GLS~\cite{Rantalankila2014}}
	    \addlegendentry{RIGOR~\cite{Humayun2014}}
	    
	    \addplot+[black,solid,mark=none, ultra thick]                          table[x=ncands,y=average_recall] {data/obj_cands/COCO_val2014_COB.txt};
        \addplot+[red,dashed,mark=triangle, mark options={solid}, mark size=1.8, thick]                          table[x=ncands,y=average_recall] {data/obj_cands/COCO_val2014_POISE.txt};
        \addplot+[blue,solid,mark=none, thick]                          table[x=ncands,y=average_recall] {data/obj_cands/COCO_val2014_MCG.txt};
        \addplot+[red,solid,mark=none, thick]                                  table[x=ncands,y=average_recall] {data/obj_cands/COCO_val2014_SCG.txt};
        \addplot+[only marks,blue,solid,mark=pentagon,mark size=1.8, thick]    table[x=ncands,y=average_recall] {data/obj_cands/COCO_val2014_GOP.txt};
	  	\addplot+[only marks,cyan,solid,mark=diamond,mark size=1.9, thick]      table[x=ncands,y=average_recall] {data/obj_cands/COCO_val2014_GLS.txt};
	    \addplot+[only marks,red,solid,mark=asterisk, mark size=1.8, thick]   table[x=ncands,y=average_recall] {data/obj_cands/COCO_val2014_RIGOR.txt};
	    \addplot+[only marks,black,solid,mark=square, mark size=1.45, thick]    table[x=ncands,y=average_recall] {data/obj_cands/COCO_val2014_SeSe.txt};
	    \addplot+[only marks,green,solid,mark=triangle, mark size=1.8, thick]    table[x=ncands,y=average_recall] {data/obj_cands/COCO_val2014_LPO.txt};
	\end{axis}
   \end{tikzpicture}}
\end{minipage}
\vspace{-2mm}
\caption{\textbf{Object proposals evaluation on PASCAL Segmentation val and MS-COCO val}: Dashed lines refer to methods that do not provide a ranked set of proposals, but they need to be reparameterized.}
\label{fig:proposals_eval}
\end{figure}

Figure~\ref{fig:proposals_eval} shows the average recall~\cite{Hosang2015} with respect to the number of object proposals.
In PASCAL Segmentation, the absolute gap of improvement of COB is at least of +13\% with the second-best technique, and consistent in all the range of number of proposals.
In MS-COCO, even though we did not train on any MS-COCO image, the percentage of absolute improvement is also consistently +13\% at least.
This shows that our contours, regions, and proposals are properly learning a generic concept of object rather than
some specific categories.

\subsection{Efficiency Analysis}
\label{sec:exp:speed}
Contour detection and image segmentation, as a preprocessing step towards high-level applications, need to 
be computationally efficient. 
The previous state-of-the-art in hierarchical image segmentation~\cite{Pont-Tuset2016,Arb+11} was of limited use in
practice due to its computational load.

As a core in our system, the forward pass of our network to compute the contour strength and 8 orientations
takes 0.28 seconds on a NVidia Titan X GPU.
Table~\ref{table:timing} shows the timing comparison between the full system COB (Ours) and some related baselines on PASCAL Context. We divide
the timing into different relevant parts, namely, the contour detection step, the Oriented Watershed Transform (OWT) and Ultrametric Contour Map (UCM) computation, and the globalization (normalized cuts) step.

\begin{table}[h]
\setlength{\tabcolsep}{4pt} 
\center
\footnotesize
\resizebox{0.96\textwidth}{!}{%
\begin{tabular}{lcccc}
\toprule
Steps          		& (1) MCG~\cite{Pont-Tuset2016} 	& (2) MCG-HED 	& (3) Fast UCMs  & (4) COB (Ours) \\
\midrule
Contour Detection 	&  \ \,3.08 				&\ \ \ 0.39*	& \ \ \ 0.39*    & \,\ 0.28*      \\
OWT and UCM   		&     11.33 				&     11.58 	& \ \,  1.63     &     0.51       \\
Globalization 		&  \ \,9.96 				&  \ \,9.97 	& \ \,  9.92     &     0.00       \\
\midrule
Total Time    		&     24.37					&     21.94		&      11.94     &  \textbf{0.79} \\
\bottomrule
\end{tabular}
}
\vspace{2mm}
\caption{\textbf{Timing experiments}: Comparing our approach to different baselines. Times computed using a GPU are marked with an asterisk.}
\label{table:timing}
\end{table}

Column (1) shows the timing for the original MCG~\cite{Pont-Tuset2016}, which uses Structured Edges (SE)~\cite{DoZi13}. As a first baseline, Column (2) displays the timing of MCG if we naively substitute SE by HED~\cite{XiTu15} at the three scales (running on a GPU).
By applying the sparse boundaries representation we reduce the UCM and OWT time from 11.58 to 1.63 seconds (Column (3)). 
Our final technique COB, in which we remove the globalization step, computes the three scales in one pass and add
contour orientations, takes 0.79 seconds in mean.
Overall, comparing to previous state-of-the-art, we get a significant improvement at a fraction of the computation time (24.37 to 0.79 seconds).

\begin{figure}[h!]
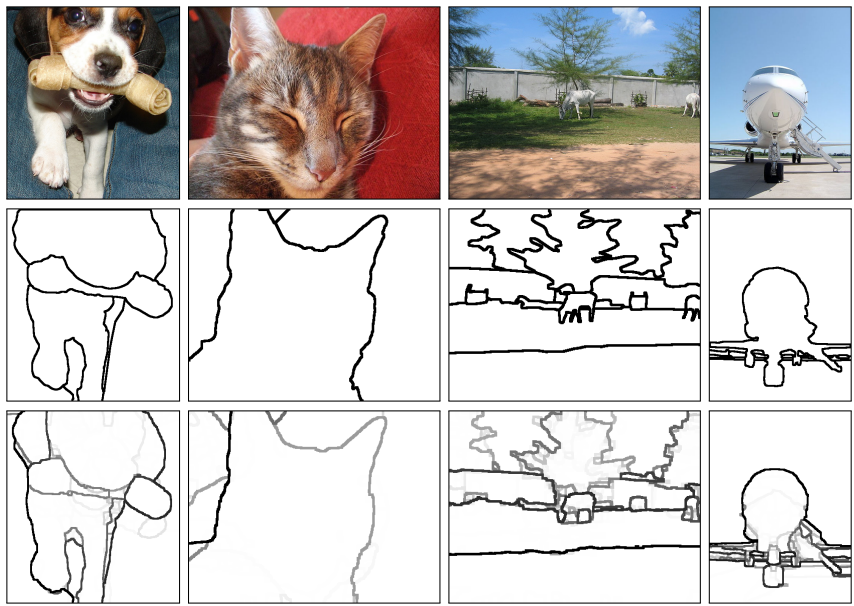

\centering
\scalebox{0.98}{\setlength{\fboxsep}{0pt}\noindent%
\showonestuff{2008_000078}{2008_000056}{2008_000009}{2008_000064}{img}{0.21}}\\[0.8mm]
\scalebox{0.98}{\setlength{\fboxsep}{0pt}\noindent%
\showonestuff{2008_000078}{2008_000056}{2008_000009}{2008_000064}{gt}{0.21}}\\[0.8mm]
\scalebox{0.98}{\setlength{\fboxsep}{0pt}\noindent%
\showonestuff{2008_000078}{2008_000056}{2008_000009}{2008_000064}{seg}{0.21}}\\
\caption{\textbf{Qualitative results on PASCAL - Hierarchical Regions}. Row 1: original images, Row 2: ground-truth boundaries, Row 3: hierarchical regions with COB.}
\label{fig:qual_cont1}
\end{figure}

\paragraph*{\textbf{Qualitative Results:}}
Figure~\ref{fig:qual_cont1} shows some qualitative results of our hierarchical contours.
Please note that COB is capable of correctly distinguishing between internal contours (e.g. cat or dog) and external, semantical, object boundaries.

\section{Conclusions}
In this work, we have developed an approach to detect contours at multiple scales, together with their orientations, in a single forward pass of a convolutional neural network. We provide a fast framework for generating region hierarchies by efficiently combining multiscale oriented contour detections, thanks to a new sparse boundary representation.
We shift from the BSDS to PASCAL in the evaluation to unwind all the potential of 
data-hungry methods such as CNNs and by observing that the performance on the BSDS is close to saturation.

Our technique achieves state-of-the-art performance by a significant margin for contour detection, the estimation of their orientation, generic image segmentation, and object proposals. We show that our architecture is modular by using two different CNN base architectures, which suggests that it will be able to transfer further improvements in CNN base architectures to perceptual grouping.
We also show that our method does not require globalization, which was a speed bottleneck in previous approaches.

All our code, CNN models, pre-computed results, dataset splits, and benchmarks are publicly available at \url{www.vision.ee.ethz.ch/~cvlsegmentation/}. 

\paragraph{\textbf{Acknowledgements:}} Research funded by the EU Framework Programme for Research and Innovation - Horizon 2020 - Grant Agreement No. 645331 - EurEyeCase. The authors gratefully acknowledge support by armasuisse, and thank NVIDIA Corporation for donating the GPUs used in this project.

\bibliographystyle{splncs}
\bibliography{0179}

\begin{thebibliography}{10}

\bibitem{Kokkinos2016}
Kokkinos, I.:
\newblock Pushing the boundaries of boundary detection using deep learning.
\newblock In: ICLR. (2016)

\bibitem{XiTu15}
Xie, S., Tu, Z.:
\newblock Holistically-nested edge detection.
\newblock In: ICCV. (2015)

\bibitem{BST15a}
Bertasius, G., Shi, J., Torresani, L.:
\newblock Deepedge: A multi-scale bifurcated deep network for top-down contour
  detection.
\newblock In: CVPR. (2015)

\bibitem{BST15b}
Bertasius, G., Shi, J., Torresani, L.:
\newblock High-for-low and low-for-high: Efficient boundary detection from deep
  object features and its applications to high-level vision.
\newblock In: ICCV. (2015)

\bibitem{She+15}
Shen, W., Wang, X., Wang, Y., Bai, X., Zhang, Z.:
\newblock Deepcontour: A deep convolutional feature learned by positive-sharing
  loss for contour detection.
\newblock In: CVPR. (2015)

\bibitem{GaLe14}
Ganin, Y., Lempitsky, V.:
\newblock N$^{4}$-fields: Neural network nearest neighbor fields for image
  transforms.
\newblock In: ACCV. (2014)

\bibitem{Krizhevsky2012}
Krizhevsky, A., Sutskever, I., Hinton, G.E.:
\newblock Imagenet classification with deep convolutional neural networks.
\newblock In: NIPS. (2012)

\bibitem{Sze+15}
Szegedy, C., Liu, W., Jia, Y., Sermanet, P., Reed, S., Anguelov, D., Erhan, D.,
  Vanhoucke, V., Rabinovich, A.:
\newblock Going deeper with convolutions.
\newblock In: CVPR. (2015)

\bibitem{SiZi15}
Simonyan, K., Zisserman, A.:
\newblock Very deep convolutional networks for large-scale image recognition.
\newblock In: ICLR. (2015)

\bibitem{He+15}
He, K., Zhang, X., Ren, S., Sun, J.:
\newblock Deep residual learning for image recognition.
\newblock In: CVPR. (2016)

\bibitem{Martin2001}
Martin, D., Fowlkes, C., Tal, D., Malik, J.:
\newblock A database of human segmented natural images and its application to
  evaluating segmentation algorithms and measuring ecological statistics.
\newblock In: ICCV. (2001)

\bibitem{Eve+12}
Everingham, M., Van~Gool, L., Williams, C.K.I., Winn, J., Zisserman, A.:
\newblock The {PASCAL} {V}isual {O}bject {C}lasses {C}hallenge 2012 {(VOC2012)}
  {R}esults.
\newblock
  http://www.pascal-network.org/challenges/VOC/voc2012/workshop/index.html

\bibitem{Har+11}
Hariharan, B., Arbel{\'a}ez, P., Bourdev, L., Maji, S., Malik, J.:
\newblock Semantic contours from inverse detectors.
\newblock In: ICCV. (2011)

\bibitem{Mot+14}
Mottaghi, R., Chen, X., Liu, X., Cho, N.G., Lee, S.W., Fidler, S., Urtasun, R.,
  Yuille, A.:
\newblock The role of context for object detection and semantic segmentation in
  the wild.
\newblock In: CVPR. (2014)

\bibitem{Arb+11}
Arbel{\'a}ez, P., Maire, M., Fowlkes, C., Malik, J.:
\newblock Contour detection and hierarchical image segmentation.
\newblock TPAMI \textbf{33}(5) (2011)  898--916

\bibitem{Pont-Tuset2016}
Pont-Tuset, J., Arbel\'{a}ez, P., Barron, J., F.Marques, Malik, J.:
\newblock Multiscale combinatorial grouping for image segmentation and object
  proposal generation.
\newblock TPAMI (2016)

\bibitem{Man+16}
Maninis, K., Pont-Tuset, J., Arbel\'{a}ez, P., Gool, L.V.:
\newblock Deep retinal image understanding.
\newblock In: MICCAI. (2016)

\bibitem{BST15c}
Bertasius, G., Shi, J., Torresani, L.:
\newblock Semantic segmentation with boundary neural fields.
\newblock In: CVPR. (2016)

\bibitem{Kho+16}
Khoreva, A., Benenson, R., Omran, M., Hein, M., Schiele, B.:
\newblock Weakly supervised object boundaries.
\newblock In: CVPR. (2016)

\bibitem{Li+16}
Li, Y., Paluri, M., Rehg, J.M., Doll{\'a}r, P.:
\newblock Unsupervised learning of edges.
\newblock In: CVPR. (2016)

\bibitem{Yan+16}
Yang, J., Price, B., Cohen, S., Lee, H., Yang, M.H.:
\newblock Object contour detection with a fully convolutional encoder-decoder
  network.
\newblock In: CVPR. (2016)

\bibitem{Najman1996}
Najman, L., Schmitt, M.:
\newblock Geodesic saliency of watershed contours and hierarchical
  segmentation.
\newblock TPAMI \textbf{18}(12) (1996)  1163--1173

\bibitem{Lee+14}
Lee, C.Y., Xie, S., Gallagher, P., Zhang, Z., Tu, Z.:
\newblock Deeply-supervised nets.
\newblock arXiv preprint arXiv:1409.5185 (2014)

\bibitem{DoZi13}
Doll{\'a}r, P., Zitnick, C.L.:
\newblock Structured forests for fast edge detection.
\newblock In: ICCV. (2013)

\bibitem{Jia+14}
Jia, Y., Shelhamer, E., Donahue, J., Karayev, S., Long, J., Girshick, R.,
  Guadarrama, S., Darrell, T.:
\newblock Caffe: Convolutional architecture for fast feature embedding.
\newblock arXiv preprint arXiv:1408.5093 (2014)

\bibitem{Martin2004}
Martin, D., Fowlkes, C., Malik, J.:
\newblock Learning to detect natural image boundaries using local brightness,
  color, and texture cues.
\newblock TPAMI \textbf{26}(5) (2004)  530--549

\bibitem{Pont-Tuset2016a}
Pont-Tuset, J., Marques, F.:
\newblock Supervised evaluation of image segmentation and object proposal
  techniques.
\newblock TPAMI \textbf{38}(7) (2016)  1465--1478

\bibitem{Zhao2015}
Zhao, Q.:
\newblock Segmenting natural images with the least effort as humans.
\newblock In: BMVC. (2015)

\bibitem{Ren2013}
Ren, Z., Shakhnarovich, G.:
\newblock Image segmentation by cascaded region agglomeration.
\newblock In: CVPR. (2013)

\bibitem{Shi2000}
Shi, J., Malik, J.:
\newblock Normalized cuts and image segmentation.
\newblock TPAMI \textbf{22}(8) (2000)

\bibitem{Felzenszwalb2004}
Felzenszwalb, P.F., Huttenlocher, D.P.:
\newblock Efficient graph-based image segmentation.
\newblock IJCV \textbf{59} (2004)  2004

\bibitem{Comaniciu2002}
Comaniciu, D., Meer, P.:
\newblock Mean shift: a robust approach toward feature space analysis.
\newblock TPAMI \textbf{24}(5) (2002)  603 --619

\bibitem{Girshick2014}
Girshick, R., Donahue, J., Darrell, T., Malik, J.:
\newblock Rich feature hierarchies for accurate object detection and semantic
  segmentation.
\newblock In: CVPR. (2014)

\bibitem{Girshick2015}
Girshick, R.:
\newblock Fast {R-CNN}.
\newblock In: ICCV. (2015)

\bibitem{Ren2015}
Ren, S., He, K., Girshick, R., Sun, J.:
\newblock Faster {R-CNN}: Towards real-time object detection with region
  proposal networks.
\newblock In: NIPS. (2015)

\bibitem{Humayun2015}
Humayun, A., Li, F., Rehg, J.M.:
\newblock The middle child problem: Revisiting parametric min-cut and seeds for
  object proposals.
\newblock In: ICCV. (2015)

\bibitem{Kraehenbuehl2015}
Kr{\"a}henb{\"u}hl, P., Koltun, V.:
\newblock Learning to propose objects.
\newblock In: CVPR. (2015)

\bibitem{Kraehenbuehl2014}
Kr{\"a}henb{\"u}hl, P., Koltun, V.:
\newblock Geodesic object proposals.
\newblock In: ECCV. (2014)

\bibitem{Uijlings2013}
Uijlings, J.R.R., van~de Sande, K.E.A., Gevers, T., Smeulders, A.W.M.:
\newblock Selective search for object recognition.
\newblock IJCV \textbf{104}(2) (2013)  154--171

\bibitem{Rantalankila2014}
Rantalankila, P., Kannala, J., Rahtu, E.:
\newblock Generating object segmentation proposals using global and local
  search.
\newblock In: CVPR. (2014)

\bibitem{Humayun2014}
Humayun, A., Li, F., Rehg, J.M.:
\newblock {RIGOR}: {R}ecycling {I}nference in {G}raph {C}uts for generating
  {O}bject {R}egions.
\newblock In: CVPR. (2014)

\bibitem{Hosang2015}
Hosang, J., Benenson, R., Doll\'ar, P., Schiele, B.:
\newblock What makes for effective detection proposals?
\newblock TPAMI \textbf{38}(4) (2016)  814--–830

\bibitem{Pont-Tuset2015b}
Pont-Tuset, J., Van~Gool, L.:
\newblock Boosting object proposals: From {Pascal} to {COCO}.
\newblock In: ICCV. (2015)

\bibitem{Lin2014a}
Lin, T., Maire, M., Belongie, S., Bourdev, L.D., Girshick, R.B., Hays, J.,
  Perona, P., Ramanan, D., Doll{\'{a}}r, P., Zitnick, C.L.:
\newblock Microsoft {COCO:} common objects in context.
\newblock {arXiv}:1405.0312 (2014)

\end{thebibliography}
\end{document}